\documentclass[journal]{IEEEtran}
\usepackage{amsmath,amsfonts}
\usepackage{algorithmic}
\usepackage{algorithm}
\usepackage{array}
\usepackage[caption=false,font=normalsize,labelfont=sf,textfont=sf]{subfig}
\usepackage{textcomp}
\usepackage{stfloats}
\usepackage{url}
\usepackage{verbatim}
\usepackage{graphicx}

\usepackage[backend=biber,style=ieee,natbib=true,maxnames=1,minnames=1]{biblatex}

\usepackage[colorlinks=true,linkcolor=blue,citecolor=blue]{hyperref}
\usepackage{url}
\usepackage{booktabs}
\usepackage{multirow}
\usepackage{algorithm}
\usepackage{algorithmic}
\usepackage{mathtools}
\usepackage{wrapfig}
\usepackage{xcolor}
\usepackage{newtxmath}
\usepackage[super]{nth}

\usepackage{ntheorem}
\theoremstyle{plain}
\theorembodyfont{\itshape}
\newtheorem{question}{Q}

\usepackage[framemethod=tikz]{mdframed}
\definecolor{rqlinecolor}{rgb}{0.122, 0.435, 0.698}
\newmdenv[innerlinewidth=0.5pt, roundcorner=4pt,linecolor=rqlinecolor,innerleftmargin=6pt,
innerrightmargin=6pt,innertopmargin=6pt,innerbottommargin=6pt]{mybox}

\usepackage{cleveref}

\newcommand{\Refsec}[1]{Section~\ref{#1}}

\newcommand{\Reffig}[1]{Figure~\ref{#1}}

\newcommand{\jax}{\textsc{jax}}
\newcommand{\brax}{\textsc{brax}}
\newcommand{\RL}{\textsc{rl}}
\newcommand{\IQM}{\textsc{iqm}}
\newcommand{\IPR}{\textsc{ipr}}

\newcommand{\MDP}{\textsc{mdp}}
\newcommand{\BPTT}{\textsc{bptt}}
\newcommand{\SAC}{\textsc{sac}}
\newcommand{\DDPG}{\textsc{ddpg}}
\newcommand{\DQN}{\textsc{dqn}}
\newcommand{\MVE}{\textsc{mve}}
\newcommand{\SVG}{\textsc{svg}}
\newcommand{\RETRACE}{\textsc{retrace}}
\newcommand{\AExpansion}{\textsc{ae}}
\newcommand{\CExpansion}{\textsc{ce}}

\newcommand{\wrt}{w.r.t.~}

\usepackage{hyperref,bookmark}
\usepackage{ifthen}

\DeclareRobustCommand{\vec}[1]{\bm{#1}}
\newcommand{\vs}{\vec{s}}
\newcommand{\va}{\vec{a}}

\newcommand{\Rsa}{\mathcal{R}(\vs,\va)}
\newcommand{\QHsa}{\textcolor{blue}{Q^H(\vec{s}, \vec{a})}}
\newcommand{\QnextHsa}{\textcolor{blue}{Q^H(\vs, \va)}}

\newcommand{\Vs}[1][]{
   \ifthenelse{ \equal{#1}{} }
      {\ensuremath{V(\vs)}}
      {\ensuremath{V^{#1}(\vs)}}
}

\newcommand{\Vsnext}[1][]{
   \ifthenelse{ \equal{#1}{} }
      {\ensuremath{V(\vs'}}
      {\ensuremath{V^{#1}(\vs')}}
}

\newcommand{\Qsa}[1][]{
   \ifthenelse{ \equal{#1}{} }
      {\ensuremath{Q(\vs,\va)}}
      {\ensuremath{Q^{#1}(\vs,\va)}}
}

\newcommand{\Qsanext}[1][]{
   \ifthenelse{ \equal{#1}{} }
      {\ensuremath{Q(\vs',\va')}}
      {\ensuremath{Q^{#1}(\vs',\va')}}
}

\usepackage{amsmath,amsfonts,bm}

\def\Figref#1{Figure~\ref{#1}}

\def\Secref#1{Section~\ref{#1}}

\def\eqref#1{equation~\ref{#1}}
\def\Eqref#1{Equation~\ref{#1}}

\def\1{\bm{1}}

\def\va{{\bm{a}}}

\def\vs{{\bm{s}}}

\DeclareMathAlphabet{\mathsfit}{\encodingdefault}{\sfdefault}{m}{sl}
\SetMathAlphabet{\mathsfit}{bold}{\encodingdefault}{\sfdefault}{bx}{n}

\newcommand{\E}{\mathbb{E}}

\DeclareMathOperator*{\argmax}{arg\,max}

\begin{document}

\title{Diminishing Return of Value Expansion Methods}

\author{Daniel Palenicek,
Michael Lutter,
Jo\~{a}o Carvalho,
Daniel Dennert,
Faran Ahmad,
and Jan Peters \IEEEmembership{Fellow, IEEE}
\IEEEcompsocitemizethanks{
\IEEEcompsocthanksitem D.~Palenicek is with the Technical University of Darmstadt, Germany, at the FG~Intelligent Autonomous Systems and hessian.AI, Germany. Email: palenicek@robot-learning.de.
\IEEEcompsocthanksitem M.~Lutter and J.~Carvalho are with the Technical University of Darmstadt, Germany, at the FG~Intelligent Autonomous Systems.
\IEEEcompsocthanksitem D.~Dennert and F.~Ahmad are M.Sc. students at the Technical University of Darmstadt, Germany.
\IEEEcompsocthanksitem J.~Peters is with the Technical University of Darmstadt, Germany at the FG~Intelligent Autonomous Systems, Germany, and hessian.AI, Germany, the German Research Center for AI (DFKI), Research Department: Systems AI for Robot Learning, the Centre for Cognitive Science at Technical University of Darmstadt, Germany, and the Robotics Institute Germany (RIG).
}
}

\markboth{SUBMITTED TO IEEE TRANSACTIONS ON PATTERN ANALYSIS AND MACHINE INTELLIGENCE}
{Palenicek \MakeLowercase{\textit{et al.}}: Diminishing Return of Value Expansion Methods}

\maketitle

\begin{abstract}
Model-based reinforcement learning aims to increase sample efficiency, but the accuracy of dynamics models and the resulting compounding errors are often seen as key limitations.
This paper empirically investigates potential sample efficiency gains from improved dynamics models in model-based value expansion methods.
Our study reveals two key findings when using oracle dynamics models to eliminate compounding errors.
First, longer rollout horizons enhance sample efficiency, but the improvements quickly diminish with each additional expansion step.
Second, increased model accuracy only marginally improves sample efficiency compared to learned models with identical horizons.
These diminishing returns in sample efficiency are particularly noteworthy when compared to model-free value expansion methods. These model-free algorithms achieve comparable performance without the computational overhead.
Our results suggest that the limitation of model-based value expansion methods cannot be attributed to model accuracy.
Although higher accuracy is beneficial, even perfect models do not provide unrivaled sample efficiency. Therefore, the bottleneck exists elsewhere.
These results challenge the common assumption that model accuracy is the primary constraint in model-based reinforcement learning.
\end{abstract}


\section{Introduction}
\IEEEPARstart{I}{nsufficient} sample efficiency is a central issue that often prevents reinforcement learning~(\RL{}) agents from learning in physical environments.
Especially in applications like robotics, samples from the real system are particularly scarce and expensive to acquire due to the high cost of operating robots.
A technique that has proven to substantially enhance sample efficiency is model-based \RL{}~\cite{deisenroth2013survey}.
In model-based \RL{}, a model of the system dynamics is usually learned from data, which is subsequently used for planning~\cite{chua2018pets,hafner2019plannet} or for policy learning~\cite{sutton1990dynaQ,janner2019mbpo}.

Over the years, the model-based \RL{} research community has identified several ways of applying (learned) dynamics models in the \RL{} framework.
The DynaQ framework, for instance, uses the model for data augmentation, where model rollouts are started from real environment states coming from a replay buffer~\cite{sutton1990dynaQ}.
Afterward, the collected data is given to a model-free \RL{} agent.
Recently, various improvements have been proposed to the original DynaQ algorithm, such as using ensemble neural network models and short rollout horizons~\cite{janner2019mbpo,lai2020bidirectional} and improving the synthetic data generation with model predictive control~\cite{morgan2021mopac}.

\citet{feinberg2018mve} proposed an alternative way of incorporating a dynamics model into the \RL{} framework.
Their model-based value expansion (\MVE{}) algorithm unrolls the dynamics model and discounts along the modeled trajectory to approximate more accurate targets for value function learning.
Subsequent works were primarily concerned with adaptively setting the rollout horizon based on some (indirect) measure of the modeling error, e.g., model uncertainty~\cite{buckmann2018steve,abbas2020selectiveMVE}, a reconstruction loss~\cite{wang2020dmve} or approximating the local model error through temporal difference learning~\cite{xiao2019adamve}.

Lastly, using backpropagation through time (\BPTT{})~\cite{werbos1990bptt}. This approach uses dynamics models in the policy improvement step to compute better policy gradients.
\citet{deisenroth2011pilco} use Gaussian process regression~\cite{rasmussen2006gp} and moment matching to find a closed-form solution for the gradient of the trajectory loss function.
Stochastic value gradients~(\SVG{})~\cite{heess2015svg} uses a model in combination with the reparametrization trick~\cite{Kingma2014VAE} to propagate gradients along real environment trajectories.
Others leverage the model's differentiability to directly differentiate through model trajectories~\cite{byravan2020ivg,amos2020sacsvg}.

On a more abstract level, \MVE{}- and \SVG{}-type algorithms are very similar.
Both learn a dynamics model and use it to generate $H$-step trajectories where $H$ is the number of modeled time steps. These trajectories are used to improve the approximated quantity of interest~---~the next-state $Q$-function in the case of \MVE{} and the $Q$-function for \SVG{}.
Value expansion methods assume that longer rollout horizons will improve learning if the model is sufficiently accurate~\citep{feinberg2018mve}.

The common opinion is that learning models with less prediction error may further improve the sample efficiency of \RL{}.
This argument is based on the fact that the single-step approximation error can become a substantial problem when using learned dynamics models for long trajectory rollouts.
Minor modeling errors can accumulate quickly when multi-step trajectories are built by bootstrapping successive model predictions.
This problem is known as the \textit{compounding model error}.
Furthermore, most bounds in model-based \RL{} are usually dependent on the model error~\cite{feinberg2018mve}. These improvement guarantees assume model errors converging to zero.
In practice, rollout horizons are often kept short to avoid significant compounding model error build-up~\cite{janner2019mbpo}, side-stepping the problem.
Intuitively, using longer model horizons has the potential to exploit the benefits of the model even more.
For this reason, immense research efforts have been put into building and learning better dynamics models.
We can differentiate between purely engineered (white-box) models and learned (black-box) models~\cite{NguyenTuong2011ModelLearningSurvey}.
White-box models offer many advantages over black-box models, as they are more interpretable and, thus, more predictable.
However, most real-world robotics problems are too complex to model analytically, and often one has to retreat to learning (black-box) dynamics models~\cite{chua2018pets,janner2020gamma}.
Recently, authors have proposed neural network models that use physics-based inductive biases, also known as grey-box models~\cite{lutter2019dln4ebc,lutter2019dln,greydanus2019hamiltonian}.

While research has focused on improving model quality, a question that has received little attention yet is: \textit{Is model-based reinforcement learning even limited by the model quality?}
For Dyna-style data augmentation algorithms, the answer is yes, as these methods treat model- and real-environment samples the same. However, for value expansion methods, i.e., \MVE{}- and \SVG{}-type algorithms, the answer is unclear. Better models would enable longer rollout horizons due to the reduced compounding model error and improve the value function approximations. However, the impact of both on the sample efficiency remains unclear.

In this paper, we empirically address this question for value expansion methods.
Using the true dynamics model, we empirically show that the sample efficiency does not increase substantially even when a perfect model is used. We find that increasing the rollout horizon with oracle dynamics as well as improving the value function approximation using an oracle model compared to a learned model at the same rollout horizon yields \textit{diminishing returns} in improving sample efficiency.
With the term \textit{diminishing~returns}, we refer to its definition in economics~\cite{case1999principles}. In the context of value expansion methods in model-based \RL{}, we mean that the marginal utility of better models on sample efficiency significantly decreases as the models improve in accuracy.

These gains in sample efficiency of model-based value expansion are especially disappointing when compared to model-free value expansion methods, e.g., \RETRACE{}~\cite{munos2016retrace}. Although model-free methods do not introduce computational overhead, performance is similar compared to model-based variants. When comparing the sample efficiency of different horizons for both model-based and model-free value expansion, one can clearly see that improvement in sample efficiency at best decreases with each additional model step along a modeled trajectory. And sometimes the overall performance even decreases for longer horizons in some environments.

\subsection{Summary of Contributions}
\noindent In this article, we empirically address the question of whether the sample efficiency of value expansion methods is limited by the dynamics model quality.
We summarize our contributions as follows.

\noindent(1) Using the oracle dynamics model, we empirically show that the sample efficiency of value expansion methods does not increase substantially over a learned model.\\

\noindent(2) We find that increasing the rollout horizon yields diminishing returns in improving sample efficiency, even with oracle dynamics.\\

\noindent(3) We show that model-free \RETRACE{} is a very strong baseline, which adds virtually no computational overhead.\\

\noindent(4) Our experiments show that for the critic expansion, the benefits of using a learned model does not appear to justify the added computational complexity.\\

This article is a substantial extension of our prior works~\cite{palenicek2023diminishing, palenicek2023revisiting}.
We extend our experiments to discrete state-action spaces, strengthening our findings.
Further, we improve the prior analysis and presentation of the results by presenting aggregated results across environments.
This strengthens both the statistical significance of our findings as well as the presentation of the results.

\subsection{Outline}
\noindent The article is structured as follows.
Section~\ref{sec:preliminaries} introduces the necessary notation and relevant background knowledge, from \RL{} fundamentals to value expansion methods.
Section~\ref{sec:problem_statement} identifies and presents the phenomenon of the diminishing returns.
Section~\ref{sec:analysis} represents the main part of this paper.
It contains the empirical evaluations and many ablations where we investigate potential reasons for the phenomenon of diminshing returns.
Finally, we draw a conclusion of the paper in Section~\ref{sec:conclusion}.

\section{Preliminaries}
\label{sec:preliminaries}

\noindent In this section, we introduce the necessary background and components of value expansion required for this paper.

\subsection{Reinforcement Learning}
\noindent Consider a discrete-time Markov Decision Process~(\MDP{})~\cite{puterman2014mdp}, defined by the tuple $\langle\mathcal{S}, \mathcal{A}, \mathcal{P}, \mathcal{R}, \rho, \gamma\rangle$
with state space $\mathcal{S}$, action space $\mathcal{A}$,
transition probability $\vs_{t+1}\sim\mathcal{P}(\:\cdot\:| \vs_t,\va_t)$,
reward function $r_t = \mathcal{R}(\vs_t,\va_t)$,
initial state distribution $\vs_0\sim \rho$
and discount factor $\gamma \in [0, 1)$.
At each time step $t$, the agent interacts with the environment according to its policy $\pi$.
Wherever it is clear, we will make use of the tick notation $\vs'$ and $\va'$ to refer to the next state and action at time $t+1$.
A trajectory ${\tau = (\vs_0, \va_0, \vs_1, \va_1, \ldots, \vs_T)}$ is a sequence of states and actions.
The objective of an \RL{} agent is to find an optimal policy $\pi^*$ that maximizes the expected sum of discounted rewards
\begin{equation*}
    \textstyle \pi^* = \argmax_\pi \mathbb{E}_{\tau\sim\rho,\pi, \mathcal{P}} \left[\sum_{t=0}^\infty \gamma^t r_t \right].
\end{equation*}
Often $\pi$ is parametrized and, in the model-free case, it is common to optimize its parameters with on-policy~\cite{sutton1999policygradienttheorem,peters2008nac,peters2010reps,schulman2015trpo,schulman2017ppo,palenicek2021survey} or off-policy gradient methods~\cite{degris2012offpac,silver2014dpg,lillicrap2015ddpg,fujimoto2018td3,haarnoja2018sac,bhatt2024crossq}.
In model-based \RL{} an approximate model $\hat{\mathcal{P}}$ of the true dynamics is learned, and optionally an approximate reward function $\hat{\mathcal{R}}$. This model is then used for data generation~\cite{sutton1990dynaQ,janner2019mbpo,cowen2022samba}, planning~\cite{chua2018pets,hafner2019plannet, lutter2021learning, lutter2021differentiable,schneider2022active} or stochastic optimization~\cite{deisenroth2011pilco,heess2015svg,clavera2020maac,amos2020sacsvg}.

\subsection{Maximum-Entropy Reinforcement Learning}
\noindent Maximum-Entropy \RL{} augments the \MDP{}'s reward~$r_t$ with a policy entropy term.
The goal is to prevent the policy from collapsing to a deterministic policy.
This implicitly enforces continuing exploration~\cite{ziebart2008maxentirl,fox2015tamingTheNoise,haarnoja2017rldeepebm,bhatt2024crossq}.
The resulting (soft) action-value function computes the expected sum of discounted soft-rewards following a policy $\pi$.
It is defined as
\begin{equation*}
    \textstyle \Qsa[\pi]=\mathbb{E}_{\pi,\mathcal{P}}\left[\sum_{t=0}^\infty \gamma^t ( r_t - \alpha \log \pi(\va_t | \vs_{t}) )\:| \vs_0=\vs,\va_0=\va\right].
\end{equation*}
The agent's goal is to maximize
\begin{equation*}
    \textstyle \mathcal{J}_\pi =
    \E_{\vs_0 \sim \rho}
    \left[ \sum_{t=0}^{\infty} \gamma^t (r_t - \alpha \log \pi(\va_t|\vs_t)) \right],
\end{equation*}
\wrt $\pi$.
\citet{haarnoja2018sac} have proposed the Soft Actor-Critic (\SAC{}) algorithm to optimize this objective. The \SAC{} actor and critic losses are defined as
\begin{align}
    \mathcal{J}_{\pi}(\mathcal{D}) &= \mathbb{E}_{\vs\sim\mathcal{D}} [-\Vs[\pi]] \nonumber \\
    &= \mathbb{E}_{\vs\sim\mathcal{D},\;\va\sim\pi(\:\cdot\:| \vs)} [  \alpha\log\pi(\va|\vs) - \Qsa[\pi] ],
    \label{eq:max_ent_actor_loss} \\
    \mathcal{J}_{Q}(\mathcal{D}) &= \mathbb{E}_{(\vs,\va,r,\vs')\sim\mathcal{D}} [(r + \gamma \Vsnext[\pi] - \Qsa[\pi])^2],\label{eq:max_ent_critic_loss}
\end{align}
where $\mathcal{D}$ is a dataset of previously collected transitions (replay buffer), and the next state value target is ${\Vsnext[\pi] = \mathbb{E}_{\va'\sim\pi(\:\cdot\:| \vs')}\left[\Qsanext[\pi]  - \alpha\log\pi(\va'|\vs') \right]}$.
Note, that $r_t + \gamma \Vsnext[\pi]$ is a target value.
Thus, we do not differentiate through it, as is common in deep \RL{}~\cite{Riedmiller2005neuralfittedq,mnih2013dqn}.

\subsection{H-step Value Expansion}
\noindent Through the recursive definition of the value- and action-value functions~\cite{Bellman1957DP}, they can be approximated by rolling out the model for $H$ time steps.
We refer to this (single sample) estimator as $H$-step value expansion, defined as
\begin{align}
    \label{eq:n_step_q}
    \QHsa = \;&\Rsa \nonumber \\
    &+ \textstyle\sum_{t=1}^{H-1} \gamma^{t} \left( r_t - \alpha \log \pi(\va_t|\vs_t)\right) \\
    &+ \textstyle\gamma^{H} V^{\pi}(\vs_H), \nonumber \\
    \text{with}\qquad\Vs[\pi] =\;&\mathbb{E}_{\va\sim\pi(\:\cdot\:| \vs)}\left[\Qsa[\pi]  - \alpha\log\pi(\va|\vs) \right], \nonumber\\
    \text{and}\quad\Qsa[0] \coloneqq\:&\Qsa[\pi]. \nonumber
\end{align}
For $H=0$ this estimator reduces to the function approximation used in \SAC{}.
It is important to note that these trajectories are on-policy.
In practice, the \SAC{} implementation uses a double $Q$-function~\cite{fujimoto2018td3,haarnoja2018sac}, but to keep the notation simple we do not include it here.
We further evaluate the expectation for the next state value function with a single sample, i.e. $\Vsnext[\pi] = \Qsanext[\pi] - \log\pi(\va'|\vs')$ with $\va'\sim\pi(\:\cdot\:| \: \vs')$, as done in~\cite{haarnoja2018sac}.
The way in which the $H$-step value expansion is used gives rise to two classes of algorithms, which we explain in the following sections.

\begin{table}[b]
    \caption{\textbf{Overview of key related literature.}
    We categorize model-based \RL{} algorithms using value expansion in the actor and/or critic update,
    and further note which ones use on-policy trajectories resulting from a dynamics model and which use off-policy trajectories from a replay buffer.
    }
    \begin{tabular}{ p{12em}p{2em}p{2em}p{4em}p{4em}  }
    \toprule
     Algorithm & \hfil\AExpansion{} & \hfil\CExpansion{} & \hfil on-policy & \hfil off-policy \\
     \midrule
     \citet{heess2015svg}~(\SVG{}) & \hfil$\times$ & & & \hfil$\times$ \\
     \citet{byravan2020ivg}~(\textsc{ivg})~ & \hfil$\times$ & & \hfil$\times$ &  \\
     \citet{amos2020sacsvg}~(\SAC{}-\SVG{}) & \hfil$\times$ & & \hfil$\times$ &  \\
     \citet{clavera2020maac}~(\textsc{maac}) & \hfil$\times$ & & \hfil$\times$ &  \\
     \citet{feinberg2018mve}~(\MVE{}) &  & \hfil$\times$ & \hfil$\times$ & \\
     \citet{buckmann2018steve}~(\textsc{steve}) &  & \hfil$\times$ & \hfil$\times$ & \\
     \citet{munos2016retrace}~(\RETRACE{}) &  & \hfil$\times$ & & \hfil$\times$ \\
     \citet{hafner2020dreamer}~(Dreamer) & \hfil$\times$  & \hfil$\times$ & \hfil$\times$ & \\
    \bottomrule
    \end{tabular}
    \label{tab:methods_overview}
\end{table}

\subsubsection{Actor Expansion}
The $H$-step value expansion can be used in the actor update by incorporating~\Eqref{eq:n_step_q} in the actor loss (\Eqref{eq:max_ent_actor_loss})
\begin{equation}
    \mathcal{J}^H_\pi(\mathcal{D}) = \mathbb{E}_{\vs\sim\mathcal{D}, \; \va\sim\pi(\:\cdot\:| \vs)} \big[ \alpha\log\pi(\va|\vs) - \QHsa \big],
    \label{eq:h_step_actor_loss}
\end{equation}
which resembles the class of \SVG{}-type algorithms~\cite{heess2015svg, byravan2020ivg,amos2020sacsvg,clavera2020maac}.
The difference to the original \SVG{} formulation is the addition of the entropy term.
Note that, for $H=0$, this definition reduces to the regular \SAC{} actor loss~(\Eqref{eq:max_ent_actor_loss}). We will refer to the $H$-step value expansion for the actor update as \textit{actor expansion}~(\AExpansion{}) from now on.

\subsubsection{Critic Expansion}
Incorporating value expansion in the policy evaluation step results in \textsc{td}-$\lambda$-style~\cite{Sutton1998,schulman2015gae} or Model-Based Value Expansion (\MVE{}) methods~\cite{wang2020dmve,buckmann2018steve}.
The corresponding critic loss is defined as
\begin{align}
    \mathcal{J}^H_{Q}(\mathcal{D}) =\;&\mathbb{E}\left[
    r + \gamma \left(\Qsanext[H] - \alpha\log\pi(\va'|\vs')\right) - \Qsa[\pi]\right]^2, \nonumber \\
    &\text{with}\quad (\vs,\va,r,\vs')\sim\mathcal{D},\;\va'\sim\pi(\:\cdot\:|  \vs').
    \label{eq:h_step_critic_loss}
\end{align}
Similar to actor expansion, for $H=0$, the critic update reduces to the \SAC{} critic update~(\Eqref{eq:max_ent_critic_loss}).
For the remainder of this paper, we will refer to the $H$-step value expansion for the critic update as \textit{critic expansion}~(\CExpansion{}).

\subsubsection{Extension to Off-Policy Trajectories}
We can naturally extend the action-value expansion in~\Eqref{eq:n_step_q} with the \RETRACE{} formulation~\cite{munos2016retrace}, which allows to use off-policy trajectories collected with a policy $\mu \neq \pi$ (e.g. stored in a replay buffer).
The $H$-step \RETRACE{} state-action value expansion (single sample) estimator is defined as
\begin{align}
    \label{eq:n_step_retrace}
    \textstyle Q^{H}_{\text{\textsc{rt}}}(\vs, \va) = \;&\Rsa \nonumber \\
    &+\textstyle \sum_{t=1}^{H-1} \gamma^{t} [ c_t r_t + c_{t-1} V(\vs_t) - c_t Q(\vs_t, \va_t) ] \\
    &+\textstyle\gamma^{H} c_{H-1} V(\vs_H), \nonumber
\end{align}
with the importance sampling weight defined as
\begin{align*}
    c_t &= \textstyle\prod_{j=1}^t \lambda\min\left(\;1,\;\frac{\pi(\va_j|\vs_j)}{\mu(\va_j|\vs_j)}\right),
    \text{ and } c_0\coloneqq 1.
\end{align*}
We set $\lambda=1$ for the remainder of this paper.
And by definition $Q^{0}_{\text{\textsc{rt}}}(\vs, \va)\coloneqq\Qsa[\pi]$.
The \RETRACE{} extension has the desirable property that  $c_t=1$ for on-policy trajectories (i.e. $\mu=\pi$).
In that case $Q^{H}_{\text{\textsc{rt}}}$ reduces to the $H$-step action-value target $Q^H$~(\Eqref{eq:n_step_q}, see Appendix~\ref{appendix:retrace_to_maxent} for a detailed derivation).
Furthermore, for $H=0$, it reduces to the \SAC{} target.
The second desirable property is that trajectories from the real environment or a dynamics model can be used.
Hence, we use the
\RETRACE{} formulation to evaluate all combinations of \CExpansion{} and \AExpansion{} with on- and off-policy data and real and model-generated data.
Since \RETRACE{} encompasses all the variants we introduced, for readability, we will from now on overload notation and define ${\QHsa \coloneqq Q^H_\text{\textsc{rt}}(\vs,\va)}$.

A natural combination is to use actor and critic expansion simultaneously.
However, we analyze each method separately to understand their individual issues better.
Table~\ref{tab:methods_overview} provides an overview of related value expansion literature.

\begin{figure*}[t]
\centering
    \includegraphics[width=\textwidth]{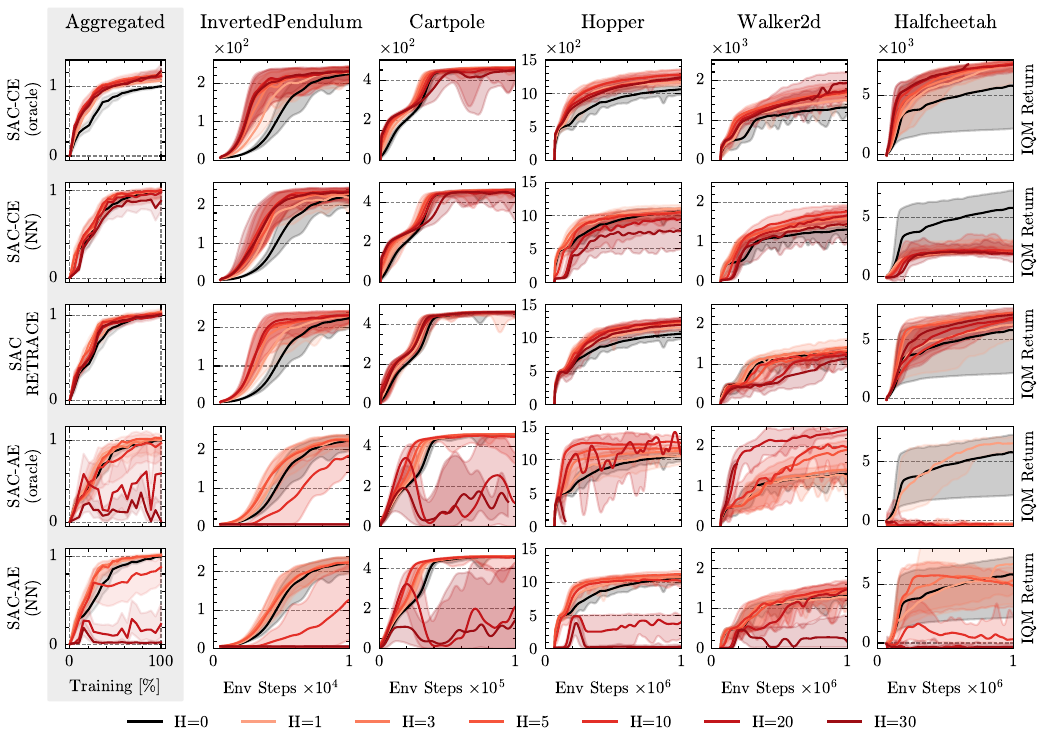}
    \caption{
    \textbf{The diminishing return on continuous control tasks.}
    Shows the diminishing return of \CExpansion{} and \AExpansion{} methods using multiple rollout horizons $H$ for a \SAC{} agent.
    \CExpansion{} does not benefit noticeably from horizons larger than 5 in most cases.
    On average, \CExpansion{} can slightly benefit from oracle dynamics over our learned model, while it is the opposite for \AExpansion{}. Looking at individual runs, oracle and learned dynamics models often perform very similarly.
    Model-free \RETRACE{} performs comparably to the learned model and even more stably with regard to longer rollout horizons.
    For \AExpansion{}, larger horizons can be detrimental, even with an oracle model.
    We plot the \IQM{} episode undiscounted return (solid line) and $90\%$ \IPR{} (shaded area) against the number of real environment interaction steps for $9$ random seeds.
    Some \AExpansion{} runs for larger $H$ terminate early due to exploding gradients (see~\Secref{sec:gradient_analysis} for detailed analysis).
    }
    \label{fig:sac_mve_ivg_retrace}
\end{figure*}

\section{The Diminishing Return of Value Expansion}
\label{sec:problem_statement}

\vspace{1em}
\begin{mybox}
\begin{question}
    \label{q:how_much}
    How much can more accurate dynamics models increase the sample efficiency of value expansion methods?
\end{question}
\end{mybox}
\noindent In this experimental section, we investigate this initial guiding research question Q1.
Intuitively, the question could be addressed by building more sophisticated models and comparing models at different levels of accuracy.
Instead of continuously attempting to develop even better models, we reverse our experimental setup and use an oracle dynamics model~---~the environment’s simulator. This model is the most accurate model that could be learned.
In this article, we attempt to analyze how value expansion methods will perform when we have more accurate dynamics models in the future.
We specifically exclude data augmentation methods, where it is trivial to realize that they can arbitrarily benefit from a perfect dynamics model. For that reason, we focus our study solely on value expansion methods.
Many of the experiments in this article are based on the idea of leveraging such an oracle dynamics model for the purpose of rigorous analysis.
In this scenario, no model error exists, which allows us to examine the impact of the compounding model error on the overall training performance or, rather, the lack thereof.
In addition to the oracle dynamics model, we present training runs with a learned neural network ensemble dynamics model and compare to them, which inevitably introduces modeling errors.
Lastly, we compare to model-free \RETRACE{}, which uses off-policy trajectories. There, the rollouts come from past behavior policies within the real environment, and therefore, \RETRACE{} does not suffer from model errors.
However, the importance sampling correction has an influence on the targets.
Comparing experiments with oracle dynamics, learned neural network model and model-free value expansion lets us investigate the next research question.

\begin{mybox}
\begin{question}
    \label{q:compounding_error}
    How does the compounding model error influence the training performance?
\end{question}
\end{mybox}

We conduct a series of empirical studies across the introduced value expansion methods on five standard continuous control benchmark tasks:
\texttt{InvertedPendulum}, \texttt{Cartpole SwingUp}, \texttt{Hopper}, \texttt{Walker2d}, and \texttt{Halfcheetah} from the \brax{}~\citep{brax2021github} physics simulator.\footnote{Code at: \href{https://github.com/danielpalen/value_expansion}{https://github.com/danielpalen/value\_expansion}}
These range from simple to complex, with the latter three including contact forces, often resulting in discontinuous dynamics.
We further provide results on four discrete state-action environments of the \texttt{MinAtar}~\citep{young19minatar} suite to strengthen our findings and extend them to discrete state-action spaces.

In the following subsections, we first present the premise of this paper~---~the two types of diminishing returns of value expansion methods~(\Refsec{sec:dim_return})~---~which originate from the research questions Q1 and Q2.
Given the oracle dynamics model, we \textit{rule out the existence of a compounding model error} and, therefore, possible impacts on the learning performance.
Based on these findings, Section~\ref{sec:analysis} defines an additional set of research questions aimed at analyzing the potential root causes of the diminishing returns.

\begin{figure}[t]
    \centering
    \includegraphics[width=\linewidth]{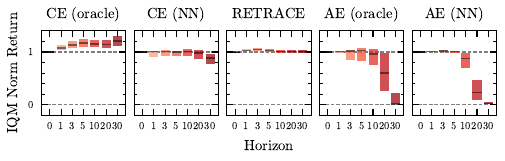}
    \caption{
    \textbf{Limited gains in final performance.}
    This figure compares the maximum performances of the different rollout horizons aggregated across environments. The results are derived from the runs from \Figref{fig:sac_mve_ivg_retrace}. We plot the \IQM{} normalized return and 95\% confidence intervals.
    Only \CExpansion{} (oracle) sees slight improvements with model rollouts, which peak at around 20\%. It is clearly visible that additional rollout steps do not contribute significantly, with the confidence intervals mostly overlapping.
    All other methods do not see any significant gains, and \AExpansion{} with long horizons again shows detrimental drops in performance.
    }
    \label{fig:aggregated_final_performance}
\end{figure}

\subsection{Diminishing Returns on Continuous Control Tasks}\label{sec:dim_return}

\noindent Each row in \Figref{fig:sac_mve_ivg_retrace} presents the learning curves of the different value expansion methods for multiple rollout horizons and model types based on a \SAC{} agent.
Each model-based value expansion method, i.e., \CExpansion{} and \AExpansion{}, is shown with an oracle and a learned model. The rollout horizons range from very short to long $H \in \{0,1,3,5,10,20,30\}$, where $H=0$ reduces to vanilla \SAC{}. Larger horizons $H \gg 30$ are not considered as the computational time complexity becomes unfeasible (see~Table~\ref{tab:computation_times} in the Appendix). In addition to comparing model-based value expansion methods, \Figref{fig:sac_mve_ivg_retrace} also includes learning curves of the model-free value expansion method \RETRACE{}.
The first column (gray background) shows aggregated performances across all five environments.
The \nth{2} to \nth{6} columns show training performances for the five continuous control environments individually.

The aggregation is based on the protocol of~\citet{agarwal2021iqm} using their \texttt{rliable} library.
With it, we aim to increase the visual clarity and the statistical significance of the numerous results across different environments.
Large parts of the paper will present this type of aggregated plots; hence, we want to explain how they are generated briefly.
To aggregate results across environments, each seed has to be normalized per environment for comparability across environments. Originally, \citet{agarwal2021iqm} used the human normalized score, as is common in \texttt{Atari}.
As there exists no human baseline for any of the tasks used in this article, we choose the maximum performance of the \textit{interquartile mean}~(\IQM{}{}) across all seeds of the \SAC{} baseline ($H=0$) in each environment, respectively.
In aggregated plots, we plot the \IQM{} and $95\%$ confidence intervals.
For individual environments, if not explicitly stated otherwise, we always plot the \IQM{} and the $90\%$ inter percentile range~(\IPR{}) over 9 random seeds.

\begin{figure}[t]
    \centering
    \includegraphics[width=\linewidth]{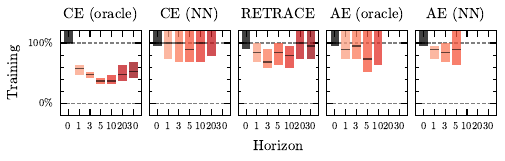}
    \caption{
    \textbf{Limited gains in learning speed.}
    This figure shows how long the different value expansion methods take to reach the maximum IQM return of the \SAC{} baseline.
    Above 100\% means that the method did not reach the performance of the \SAC{} baseline within the considered timeframe.
    We see that \CExpansion{} (oracle) benefit in terms of the learning speed.
    However, this benefit quickly diminishes with increasing rollout horizons.
    All other methods do not benefit considerably, with \RETRACE{} being marginally faster than \CExpansion{}~(NN).
    \AExpansion{} methods again show detrimental drops in performance for larger horizons.
    }
    \label{fig:aggregated_learning_speed}
\end{figure}

\subsubsection{Diminishing Return of Rollout Horizons}
The aggregated results in \Figref{fig:sac_mve_ivg_retrace} reveal that within short horizons, less than $5$ steps, we consistently observe that a single increase in the rollout horizon leads to increased performance in most cases.
However, the rate of improvement quickly decreases with each additional expansion step.
All variants of \SAC{} reach their peak performance at most $H=5$.
For more than $5$ steps, the sample efficiency and obtained reward do not increase anymore and can even decrease with slight per-environment variations.
We refer to this phenomenon as the \textit{diminishing return} of rollout horizon.
\SAC{}-\CExpansion{} (\nth{1} \& \nth{2} rows) performs better than \SAC{}-\AExpansion{} (\nth{4} \& \nth{5} rows) for both the oracle and the learned dynamics models. While for shorter horizons, the performance is comparable, the performance of \SAC{}-\AExpansion{} degrades significantly for longer horizons. For \SAC{}-\AExpansion{}, learning becomes unstable, and no good policy can be learned for most systems using longer horizons.
Thus, increasing rollout horizons does not yield unrivaled sample efficiency gains. As diminishing returns with respect to the rollout horizon are observed for both the oracle and the learned dynamics model, the reduced performance of the longer rollout horizon cannot be explained by the compounding model errors as hypothesized by prior work~\cite{feinberg2018mve}.

Figures~\ref{fig:aggregated_final_performance} \&~\ref{fig:aggregated_learning_speed} provide another view on this phenomenon by evaluating two different metrics taken from the same experiments.
\Figref{fig:aggregated_final_performance} compares the maximum performances of the different rollout horizons, aggregated across environments.
\Figref{fig:aggregated_learning_speed} compares learning speeds of different rollout horizons. It shows how long each horizon takes to reach the maximum performance of the \SAC{} baseline.
In this figure, above 100\% means that the performance of the baseline was not reached within the considered timeframe.
Both figures show that even with oracle dynamics, \CExpansion{} shows only slight gains in its maximum performance and some speedup in learning speed.
Horizons larger than $5$ do not significantly increase performance in either metric.
Every other expansion method does not benefit considerably.
In the case of \AExpansion{}, longer rollout horizons are detrimental in terms of learning performance.

\subsubsection{Diminishing Return of Model Accuracy}
When directly comparing the performance of the oracle and learned dynamics models in the \nth{1} and \nth{2} rows of \Figref{fig:sac_mve_ivg_retrace}, often the individual learning curves are very similar.
The performance of the learned model is often only slightly reduced compared to the oracle.
However, the degraded performance is barely visible for most systems when comparing the learning curves side by side.
This trend can be observed across most environments, algorithms, and horizons.

The aggregated results in Figures~\ref{fig:sac_mve_ivg_retrace},~\ref{fig:aggregated_final_performance}~\&~\ref{fig:aggregated_learning_speed} show more clearly that on average \SAC{}-\CExpansion{} benefits most from an oracle dynamics model reaching up to $\sim20\%$ higher final performance compared to the \SAC{} baseline and the neural network model.
On the contrary, \SAC-\AExpansion{} does not benefit from oracle dynamics with regard to final performance. In fact, the neural network model learns slightly faster.
Therefore, the reduced model accuracy and the subsequent compounding model error do not appear to significantly and consistently impact the sample efficiency of the presented value expansion methods.
We conclude that one cannot expect large gains in sample efficiency with value expansion methods across the board by further investing in developing more accurate dynamics models.
This is not to say that poor dynamics models do not harm the learning performance of value expansion methods.
Rather, we observe that overcoming the small errors of current models towards oracle dynamics will, at best, result in small improvements for value expansion methods on the continuous control tasks studied.
We argue that putting these slight performance gains into perspective, they are underwhelming.
Given the significant increase in required compute and the difficulty of learning an oracle dynamics model, it is questionable whether these marginal performance gains justify the added complexity and cost.

\subsubsection{Model-free Value Expansion}
When comparing the model-based value expansion methods to the model-free \RETRACE{} (\Figref{fig:sac_mve_ivg_retrace}, \nth{3} row), the sample efficiency of the learned model-based variant is nearly identical. For long horizons, when aggregating the results across environments, it is even slightly worse than the model-free counterpart.
On average, \RETRACE{} shows very stable learning performance across all horizons.
Figures~\ref{fig:aggregated_final_performance}~\&~\ref{fig:aggregated_learning_speed} also nicely illustrate \RETRACE{}'s learning stability.
They also show marginal gains in learning speed for short horizons and none for long horizons.
Studying the environments individually, similar to the model-based variants, \RETRACE{} increases the sample efficiency for shorter horizons.
For longer horizons, in some cases, it can even result in decreasing performance.
This minor difference between both approaches becomes especially surprising as the model-based value expansion methods introduce a large computational overhead.
While the model-free variant only utilizes the collected experience, the model-based variant requires learning a parametric model and unrolling this model at each update of the critic and the actor.
Empirically, the model-free variant is up to $15\times$ faster in wall-clock time compared to the model-based \CExpansion{} at $H=5$.
Therefore, model-free value expansion methods are a very strong baseline to model-based value expansion methods, especially when considering the computational overhead.

\begin{figure}[t]
    \centering
    \includegraphics[width=\linewidth]{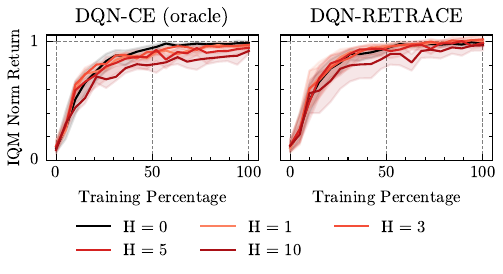}
    \caption{\textbf{Diminishing return on \texttt{MinAtar} discrete state-action environments.}
    We show \IQM{} normalized returns aggregated over four environments and 9 seeds, each with $95\%$ confidence intervals.
    \DQN{} does not benefit from value expansion even with oracle dynamics. Long rollout horizons can even hurt performance.}
    \label{fig:discrete_aggregated}
\end{figure}

\subsection{Diminishing Returns on Discrete State-Action Tasks}
\noindent To strengthen our findings with regard to the identified diminishing returns, we run additional experiments on four discrete state-action environments of the \texttt{MinAtar} training suite~\citep{young19minatar}: \texttt{Breakout}, \texttt{Freeway}, \texttt{Asterix} und \texttt{Space Invaders}.
We chose \texttt{MinAtar} as it implements miniaturized environments of the classic \texttt{Atari 2600} games in \jax{}, which is a common benchmark for discrete \RL{}.
For our experiments with oracle dynamics, it is important that the environments are implemented in \jax{}, such that they can be run on the GPU and integrated with the agent code directly to optimize resource utilization.
As a base algorithm, we use \DQN{}~\cite{mnih2013dqn} with the $H$-step Q-function analogous to before (\Eqref{eq:n_step_q}).
\DQN{} only requires learning a Q-function and no parametric policy, compared to \SAC{}, as for small, discrete action spaces, it is computationally feasible to evaluate it entirely.
Therefore, the policy is simply defined as
$
    \textstyle \pi(\vs) = \arg\max_\va Q(\vs,\va).
$
As there is no policy improvement step, in the discrete state-action case, we only investigate \CExpansion{}.

Figure~\ref{fig:discrete_aggregated} shows \DQN{}-\CExpansion{} with oracle dynamics and \DQN{}-\RETRACE{} results on \texttt{MinAtar} for $H\in\{0,1,3,5,10\}$.
Each curve is aggregated over all four environments with 9 seeds each.
Neither \DQN{}-\CExpansion{}, nor, \DQN{}-\RETRACE{}, benefit from longer rollout horizons, even with oracle dynamics, showing the same kinds of diminishing returns.

For the remainder of the paper, we will return to the continuous case and analyze this setting in more depth.

\section{Analysis of the Diminishing Return}
\label{sec:analysis}
\noindent After identifying the diminishing returns we now want to investigate potential causes.
To guide our analysis in this section, we formulate the following research questions.

\begin{mybox}
\begin{question}
    How do different rollout horizons influence the Q-function targets that are being generated?
\end{question}
\begin{question}
    Can increasing variance in the generated Q-function target values be a cause of diminishing returns?
\end{question}
\begin{question}
    How do longer rollouts influence the gradients?
\end{question}
\begin{question}
    Is the discount factor a cause for diminishing returns?
\end{question}
\end{mybox}

Each of the following subsections is dedicated to the investigation of one of the above research questions.

\begin{figure*}[t]
    \centering
    \includegraphics[width=\linewidth]{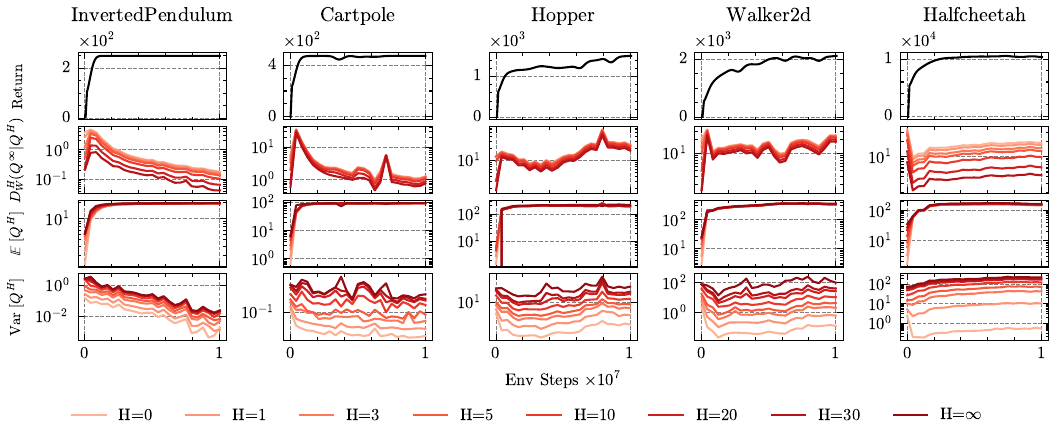}
    \caption{
    \textbf{Analysis of $H$-step targets} for multiple checkpoints along the training process.
    The \nth{1} row shows the undiscounted episodic return of \SAC{} along training for one seed. The checkpoints along this single training run are used for the analyses in the following rows.
    The \nth{2} row shows the Wasserstein distance between the sample-based target distribution predicted by the $H$-step estimator and the true target distribution represented by Monte Carlo samples.
    \nth{3} and \nth{4} rows display the mean and variance over particles of the sample-based target distributions.
    The mean targets for different horizons are roughly equal, while the variance increases with the horizon.
    }
    \label{fig:sac_h_target_wasserstein}
\end{figure*}

\subsection*{\normalfont{\textbf{Q 3}}\textit{\quad $Q$-Target Distributions}}
\label{sec:target_distributions}
\noindent Estimating $\QHsa$ by rolling out a model with a stochastic policy,
can be seen as traversing a Markov reward process for $H$ steps.
It is well known that here the target estimator's variance increases linearly with $H$, similar to the \textsc{reinforce} policy gradient algorithm~\cite{Pflug1996OptimizationStochasticModels,carvalho2021ijcnn}.
At the same time, the value expansion targets with oracle dynamics should model the true return distribution better.
In order to confirm this, we compare the true return distribution, approximated by particles of infinite horizon Monte Carlo returns, to the targets produced by different rollout horizons using
$$
    D_{\text{W}}^H = \mathbb{E}_{\vs,\va\sim\mathcal{D}}\big[D_{\text{W}}\big(\big\{Q^\infty_p(\vs,\va)\big\}_{p=1 \ldots P}\big|\big\{Q^H_p(\vs,\va)\big\}_{p=1 \ldots P}\big)\big],
$$
where $D_W$ is the Wasserstein distance~\cite{villani2009optimal}, $P=100$ particles per target (to model the distribution over targets).
$Q^\infty(\vs,\va)$ is the true Monte Carlo return to compare against~(for computational reasons we use a rollout horizon of 300 steps).
Finally, we evaluate the expectation of this estimator over $10^4$ samples $(\vs,\va)\sim\mathcal{D}$ drawn from the replay buffer.
We repeat this analysis for all horizons and $25$ checkpoints along the vanilla \SAC{} training.
Figure~\ref{fig:sac_h_target_wasserstein} \nth{2} row shows that with increasing horizons $D_{\text{W}}^H$
decreases, i.e., as expected, longer rollouts do indeed capture the true return distribution better.
The \nth{3} and \nth{4} row of \Figref{fig:sac_h_target_wasserstein} depict
the expectation of the target values particles over a replay buffer batch and the corresponding variance
\begin{align*}
    \mathbb{E}_{\vs,\va\sim\mathcal{D}}[&\mathbb{E}[\{Q^H_p(\vs,\va)\}_{p=1..P}]], \\
    \mathbb{E}_{\vs,\va\sim\mathcal{D}}[&\text{Var}[\{Q^H_p(\vs,\va)\}_{p=1..P}]].
\end{align*}
The expected values of the target distribution $Q^H$ are very similar for different horizons.
This observation means that the approximated $Q^0$-function already captures the true return very well and, therefore, a model-based expansion only marginally improves the targets in expectation.
The \nth{4} row shows, however, that the variance over targets increases by orders of magnitude with the increasing horizon.

\begin{figure}[t]
\centering
    \includegraphics[width=\linewidth]{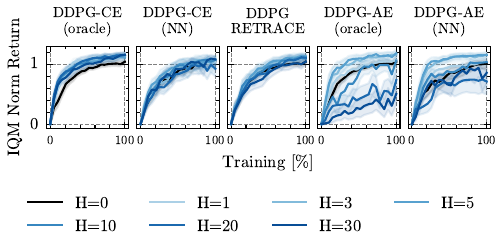}
    \caption{
    \textbf{Deterministic policy experiments.}
    Similar to~\Figref{fig:sac_mve_ivg_retrace}, this plot shows the diminishing return of \CExpansion{} and \AExpansion{} for a \DDPG{} agent in aggregated form.
    We notice the same behaviors where \DDPG{}-\CExpansion{} does not benefit noticeably from larger horizons in most cases.
    Again, learned neural network models and model-free \RETRACE{} performance are comparable.
    For \DDPG-\AExpansion{}, larger horizons can be detrimental, even with an oracle model.
    We plot the \IQM{} norm return (solid line) and $95\%$ confidence intervals (shaded area) against the number of real environment steps for $9$ random seeds.
    }
    \label{fig:ddpg_aggregated}
\end{figure}

\subsection*{\normalfont{\textbf{Q 4}}\textit{\quad Increasing Variance of $\QHsa$}}
\label{sec:target_analysis}
\noindent We empirically observed the increase in variance of the $Q$-function targets when increasing the horizon. To test whether the increased variance of the value expansion targets explains the diminishing returns, we conduct the following two experiments.
First, we rerun the experiments from~\Secref{sec:dim_return} based on \DDPG{}~\cite{lillicrap2015ddpg} instead of \SAC{}. As \DDPG{} uses a deterministic policy and the oracle dynamics are deterministic as well, increasing the horizon does not increase the variance of the $Q$-function targets. Second, we employ variance reduction when computing the targets with \SAC{} by averaging over multiple particles per target.

\begin{figure*}[b]
    \centering
    \includegraphics[width=\linewidth]{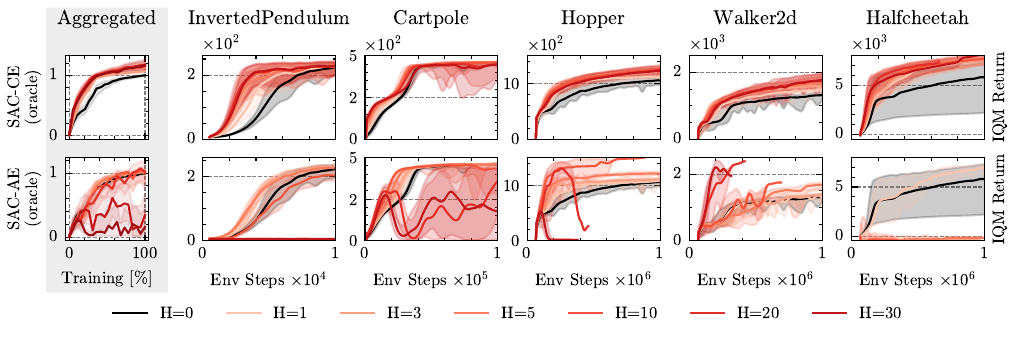}
    \caption{
    \textbf{Variance reduction through particles} for \SAC{}-\CExpansion{} and \SAC{}-\AExpansion{} experiments.
    Each $Q$-function target is averaged over multiple particles to reduce the variance (30 particles for \SAC{}-\CExpansion{} and 10 for \SAC{}-\AExpansion{}).
    Compared to experiments from~\Figref{fig:sac_mve_ivg_retrace}, we see that variance reduction gives the same qualitative training performance.
    Therefore, the increased variance of the $Q$-function targets is not the root cause of the diminishing returns in value expansion.
    }
    \label{fig:sac_particles}
\end{figure*}

\subsubsection*{Deterministic Policies}
\Figref{fig:ddpg_aggregated} shows the aggregated learning curves of \DDPG{} for the same systems and horizons as \Figref{fig:sac_mve_ivg_retrace}.
We provide individual learning curves in \Figref{fig:ddpg_all} in the Appendix.
Since both the oracle dynamics and the policy are deterministic, increasing the horizon does not increase the variance of the $Q$-function targets. \DDPG{} shows the same diminishing returns as \SAC{}. When increasing the horizon, the gain in sample efficiency decreases with each additional step. Longer horizons are not consistently better than shorter horizons. For longer horizons, the obtained reward frequently collapses, similar to \SAC{}. Therefore, the increasing variance in the $Q$-function targets cannot explain the diminishing returns.

\subsubsection*{Variance Reduction Through Particles}
To reduce the $Q$-function target variance when using \SAC{}, we average the computed target $\QHsa$ over multiple particles per target estimation. For \SAC{}-\CExpansion{}, we used 30 particles and 10 particles for \SAC{}-\AExpansion{}. The corresponding learning curves are shown in~\Figref{fig:sac_particles}.
Comparing the experiments with variance reduction with the prior results from~\Figref{fig:sac_mve_ivg_retrace}, there is no qualitative difference in training performance. Increasing the horizons by one additional step does not necessarily yield better sample efficiency. As in the previous experiments, for longer horizons, the learning curves frequently collapse, and the performance deteriorates even with the oracle dynamics model and averaging over multiple particles. Therefore, the increasing variance of the \SAC{} $Q$-function targets does not explain
the diminishing returns of value expansion methods.

In conclusion, both deterministic policies as well as variance reduction techniques show that the increasing variance of $\QHsa$ does not explain the diminishing returns of value expansion methods.

\subsection*{\normalfont{\textbf{Q 5}}\textit{\quad Impact on the Gradients}}
\label{sec:gradient_analysis}
\noindent Next, we analyze the gradient statistics during training to observe whether the gradient deteriorates for longer horizons.
The gradient standard deviation for a subset of environments and different horizons is shown in~\Figref{fig:expansion_selected_grads}.
In the Appendix, the gradient mean and standard deviation is shown for all environments, including both \SAC{} and \DDPG{}. See Figures~\ref{fig:critic_grads_all}~\&~\ref{fig:actor_grads_all} for all \SAC{} experiments and Figures~\ref{fig:critic_grads_all_ddpg}~\&~\ref{fig:actor_grads_all_ddpg} for the \DDPG{} experiments.

\begin{figure}[t]
    \centering
    \includegraphics[width=\linewidth]{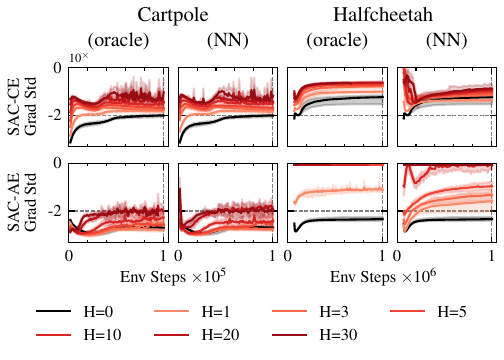}
\caption{
\textbf{Critic and actor gradient's standard deviation} for Cartpole and Halfcheetah and different rollout horizons with both an oracle dynamics model and a learned dynamics model. (a) We observe that the \SAC{}-\CExpansion{} critic gradients show low variance.
(b) We see that the \SAC{}-\AExpansion{} actor gradient variance explodes for Halfcheetah with longer horizons.
In this Figure, we clip values to $10^0$, which becomes necessary for visualizing longer horizons on the Halfcheetah due to exploding gradients.
}
\label{fig:expansion_selected_grads}
\end{figure}

\subsubsection*{Critic Expansion Gradients}
The \nth{1} row of Figure~\ref{fig:expansion_selected_grads} shows the \SAC{}-\CExpansion{} critic gradient's standard deviation (averaged over a replay buffer batch), over all dimensions of the $Q$-function parameter space, along the training procedure of \SAC{}-\CExpansion{}.
For all environments and horizons, the gradient mean and standard deviation appear within reasonable orders of magnitude (also compare~\Figref{fig:critic_grads_all} in the Appendix).
Although, as expected, we observe a slight increase in the standard deviation with an increasing horizon for both the oracle and the learned neural network model.
The mean is around zero $(10^{-5})$ and the standard deviation between $10^{-3}$ and $10^{-1}$.
Therefore, for both \SAC{}-\CExpansion{} and \DDPG{}-\CExpansion{}, there is no clear signal in the gradient's mean or standard deviation that links to the diminishing returns of \CExpansion{} methods.
The little difference w.r.t. horizon length is expected because the critic gradient $\nabla_Q \mathcal{J}^H_Q$ of the loss~(\Eqref{eq:h_step_critic_loss}) is only differentiating through $Q^{\pi}$ in the squared loss since $\QnextHsa$ is assumed to be a fixed target, a common assumption in most \RL{} algorithms~\cite{Ernst2005batchrl,Riedmiller2005neuralfittedq,mnih2013dqn}.

\subsubsection*{Actor Expansion Gradients}
The second row of Figure~\ref{fig:expansion_selected_grads} shows the actor gradient's standard deviation (averaged over a replay buffer batch) over all dimensions of the policy parameter space,
along the training process for \SAC{}-\AExpansion{}.
Contrary to \CExpansion{}, \AExpansion{} backpropagates through the whole rollout trajectory to compute the gradient \wrt the policy parameters.
This type of recursive computational graph is well known to be susceptible to vanishing or exploding gradients~\cite{pascanu2013difficulty}.
Comparing rows four and five of \Reffig{fig:sac_mve_ivg_retrace}, we observe that a better dynamics model does not directly translate into faster learning or higher rewards since both the oracle and neural network model show the same effect~---~lower returns for larger horizons.
For \SAC{}-\AExpansion{}, we partially correlate these results with the actor gradients in Figure~\ref{fig:expansion_selected_grads}.
For example, the standard deviation of the Halfcheetah actor gradients explodes for horizons larger or equal to $3$ for the oracle and $20$ for the learned models.
This problem of exploding gradients can also be seen in the Hopper environment using learned dynamics (Appendix Figure~\ref{fig:actor_grads_all}).
This increase in variance can be correlated with the lower returns as \SAC{}-\AExpansion{} frequently fails to solve the tasks for long horizons. A similar phenomenon can be seen in the \DDPG{}-\AExpansion{} experiments as well, where the policy and oracle dynamics are deterministic.
Therefore, increasing the horizon must be done with care even if a better model does not suffer from compounding model errors.
For \AExpansion{}, building a very accurate differentiable model that predicts into the future might not be worth it due to gradient instability. For \AExpansion{}, it might only be possible to expand for a few steps before requiring additional methods to handle gradient variance~\cite{parmas2018pipps}.

\subsection*{\normalfont{\textbf{Q 6}}\textit{\quad Analysis of Discount Factor}}
\begin{figure}[t]
    \centering
    \includegraphics[width=\linewidth]{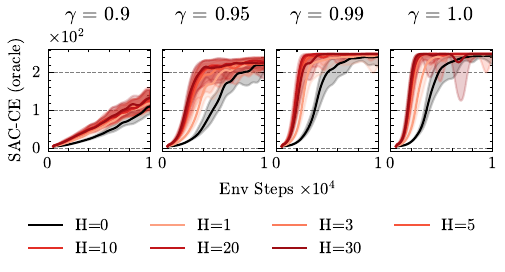}
    \caption{
    \textbf{Ablation of discount factors $\gamma$} on the InvertedPendulum environment.
    The discount factor does not appear to influence the diminishing returns, and therefore, the discount factor is not the source of diminishing returns.}
    \label{fig:different_discount}
\end{figure}

\noindent Another hypothesis for the source of the diminishing return could be the discount factor~$\gamma$.
\Figref{fig:different_discount} shows an ablation study over different discount factors on the InvertedPendulum environment.
The phenomenon of diminishing returns is obvious in all experiments, regardless of the discount factor. We do not see an influence discount factor~$\gamma$ on the diminishing return in this experiment.
We conclude that the discount factor is not the source of diminishing returns.

\section{Conclusion}
\label{sec:conclusion}

\noindent In this paper, we answer the question of whether dynamics model errors are the main limitation for improving the sample efficiency of model-based value expansion methods.
Our evaluations focus on value expansion methods as model-based data augmentation approaches (i.e. Dyna-style \RL{} algorithms) clearly benefit from a perfect model.
By using an oracle dynamics model and thus avoiding model errors altogether, our experiments show that the benefits of better dynamics models for value-expansion methods yield diminishing returns.
We show these diminishing returns for the case of infinite horizon, continuous state- and action space problems with deterministic dynamics and for discrete state-action spaces.
More specifically,
(1)
we observe the \textit{diminishing return of model accuracy}, where more accurate dynamics models do not significantly increase the sample efficiency of model-based value expansion methods.
And
(2) the \textit{diminishing return of rollout horizon},
where icreasing the rollout horizons by an additional step yields decreasing gains in sample efficiency at best.
These gains quickly plateau for longer horizons, even for oracle models that do not have compounding model errors.
For some environments, longer horizons even hurt sample efficiency.
Furthermore, we observe that model-free value expansion delivers on-par performance compared to model-based methods without the added computational overhead.
Therefore, these model-free
methods are a very strong baseline with lower computational cost.
Overall, our results suggest that the computational overhead of model-based value expansion methods often outweighs the benefits
compared to model-free ones, especially when computational resources are limited.

\section*{Acknowledgments}
\noindent This work was funded by the Hessian Ministry of Science and the Arts (HMWK) through the projects ``The Third Wave of Artificial Intelligence - 3AI'' and hessian.AI.
Calculations for this research were conducted on the Lichtenberg high-performance computer of the TU Darmstadt.
Jo\~{a}o Carvalho is funded by the German Federal Ministry of Education and Research (project IKIDA, 01IS20045).

\printbibliography

\section*{Biography Section}
\begin{IEEEbiography}[{\includegraphics[width=1in,height=1.25in,clip,keepaspectratio]{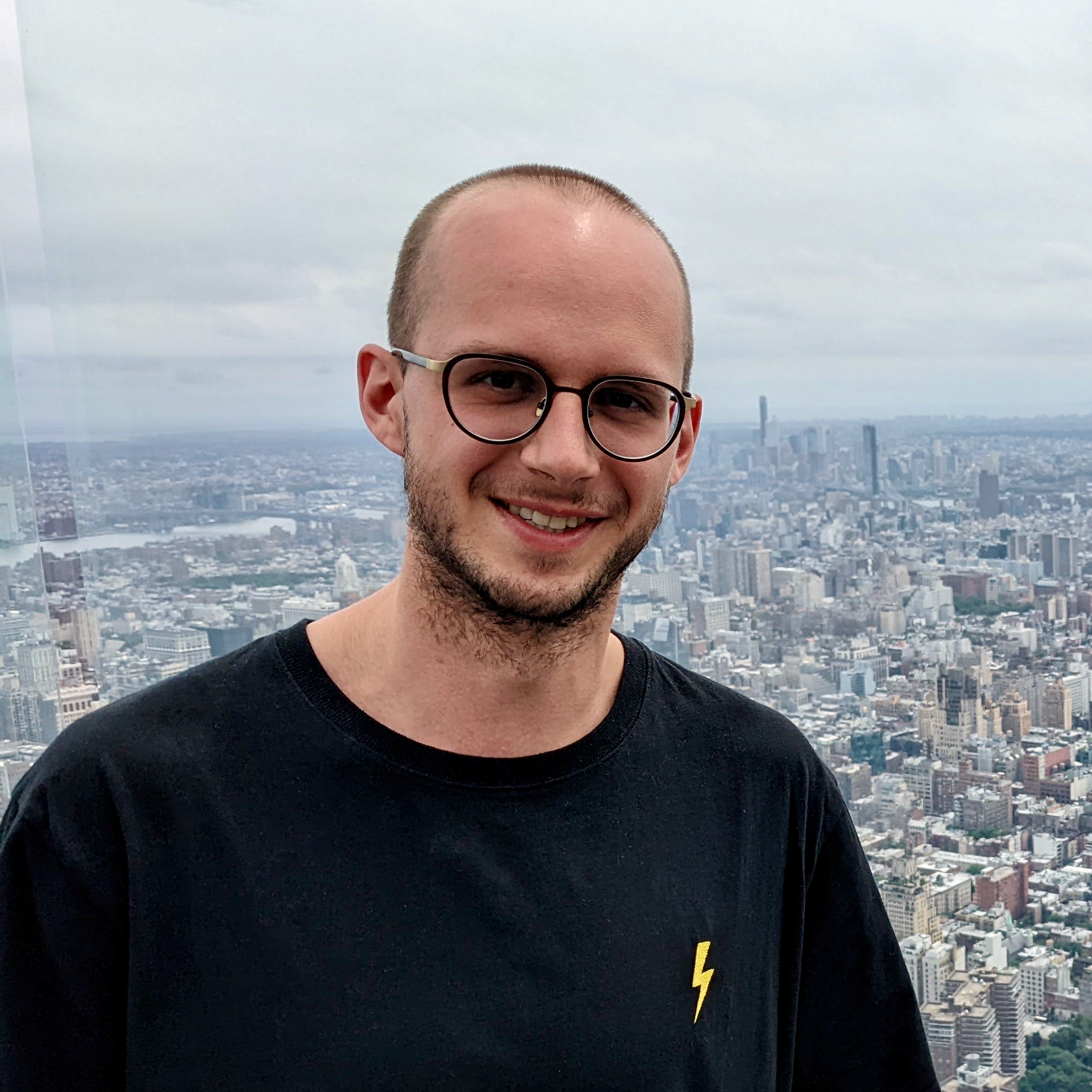}}]{Daniel Palenicek} is a Ph.D. student with Prof. Jan Peters at the Institute for Intelligent Autonomous Systems~(IAS) Technical University of Darmstadt since October 2021. During his Ph.D., Daniel works on the 3AI project at hessian.AI and focuses on improving the learning performance of reinforcement learning agents.
Before, Daniel received his M.Sc. and B.Sc. in Information Systems from the Technical University of Darmstadt.
During his M.Sc., he completed two research internships at the Bosch Center for Artificial Intelligence and Huawei R\&D London.
\end{IEEEbiography}
\begin{IEEEbiography}[{\includegraphics[width=1in,height=1.25in,clip,keepaspectratio]{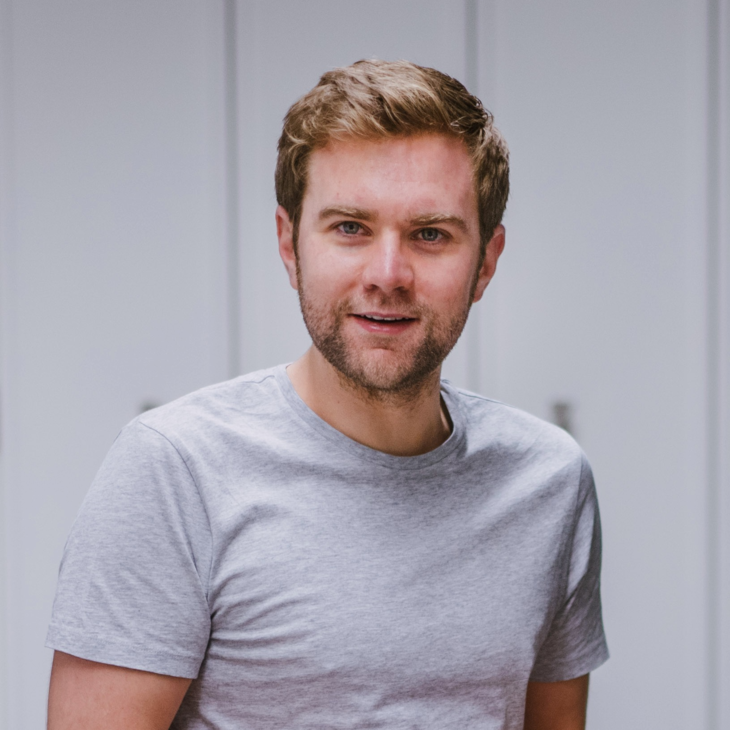}}]{Michael Lutter} completed his Ph.D. at the Institute for Intelligent Autonomous Systems~(IAS) TU Darmstadt in 2022. His thesis on "Inductive Biases in Machine Learning for Robotics and Control" was published in in the Springer STAR series. Michael has received multiple awards including the George Giralt Award for the best European robotics Ph.D. thesis in 2022 as well as the German AI Newcomer award in 2019.
\end{IEEEbiography}
\begin{IEEEbiography}[{\includegraphics[width=1in,height=1.25in,clip,keepaspectratio]{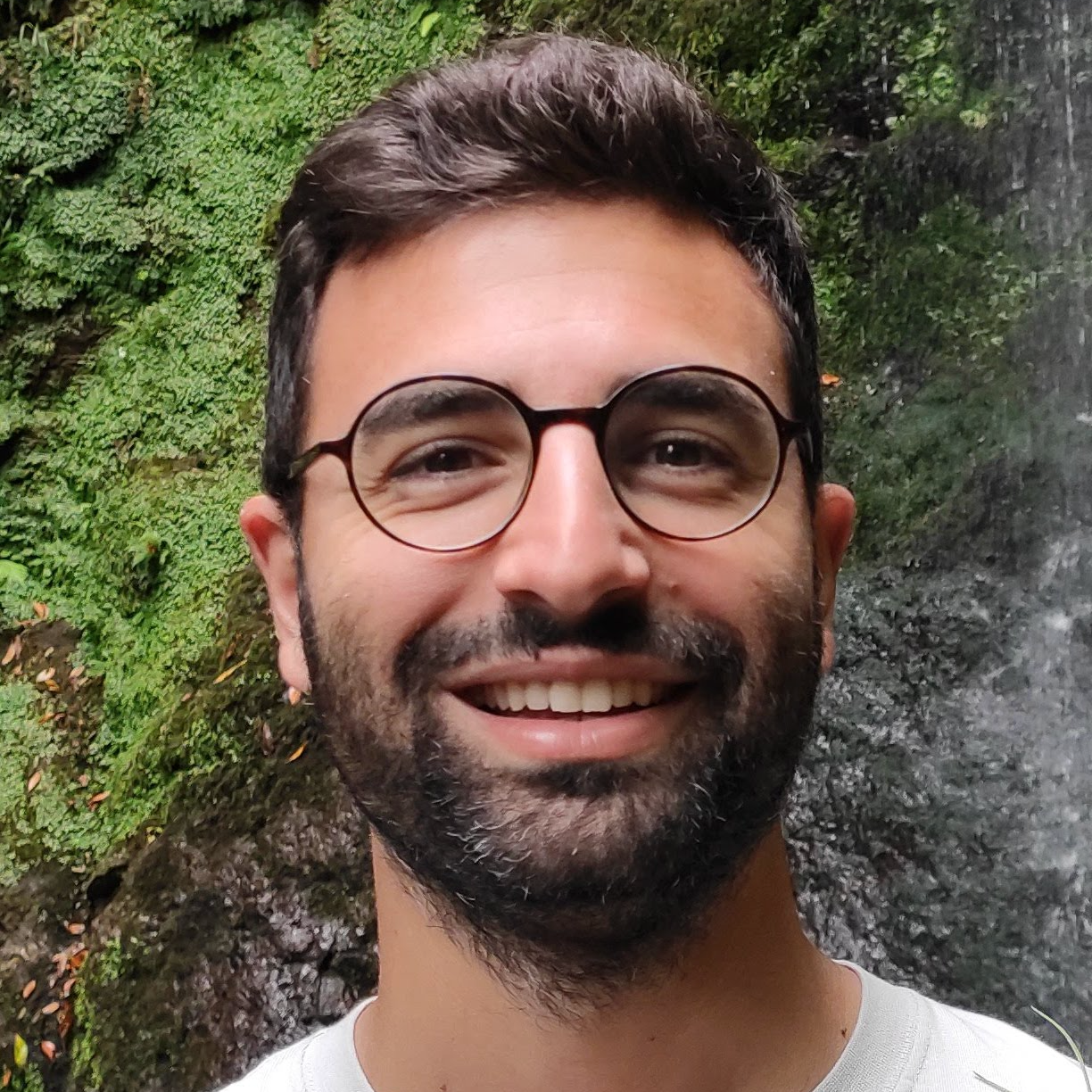}}]{Jo\~{a}o Carvalho}
is a Ph.D. student with Prof. Jan Peters at the Institute for Intelligent Autonomous Systems~(IAS) Technical University of Darmstadt.
Previously, he completed a M.Sc. degree in Computer Science from the Albert-Ludwigs-Universit\"at Freiburg and studied Electrical and Computer Engineering at the Instituto Superior Técnico of the University of Lisbon. His research is focused on learning algorithms for control and robotics.
\end{IEEEbiography}
\begin{IEEEbiography}[{\includegraphics[width=1in,height=1.25in,clip,keepaspectratio]{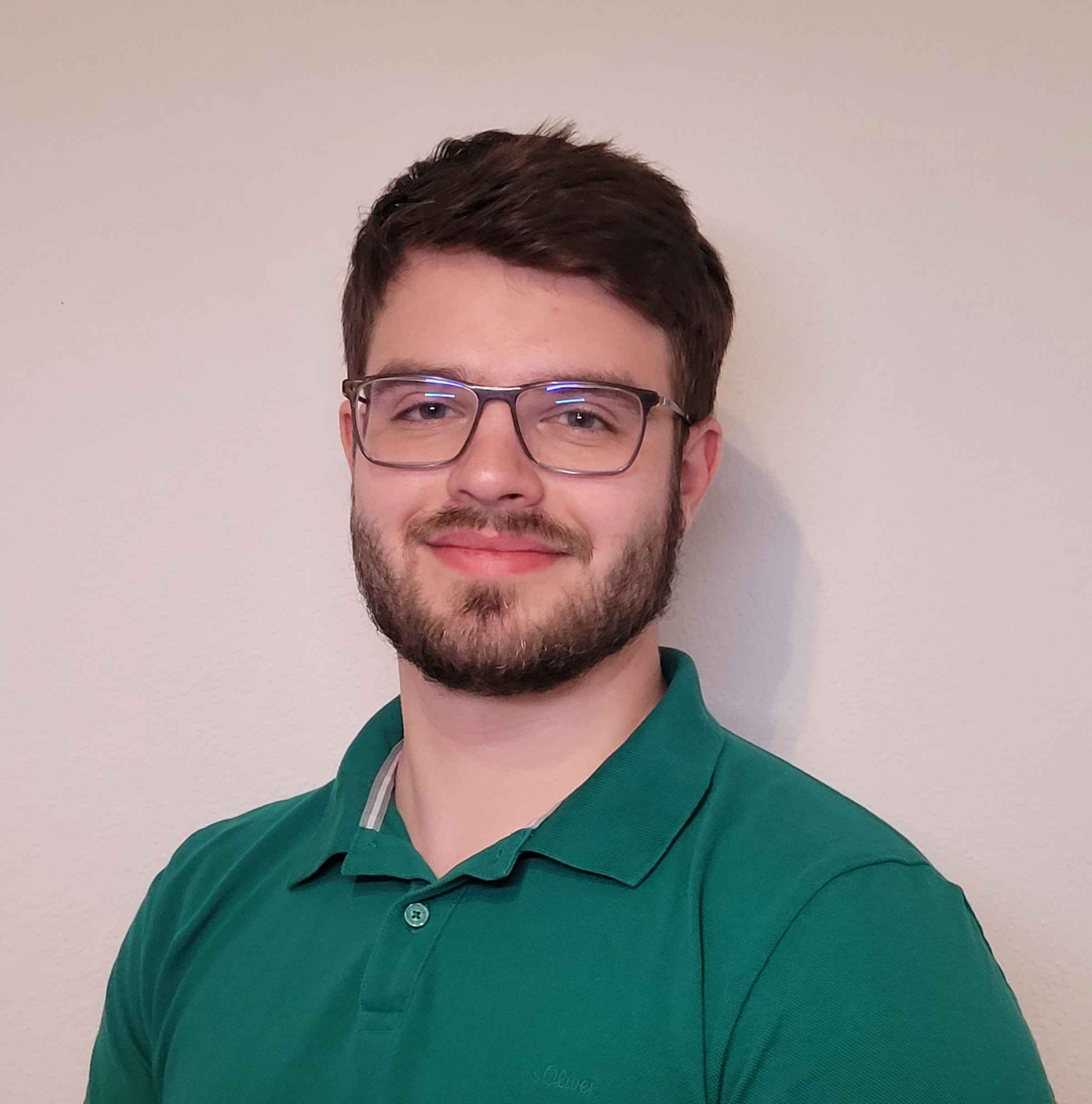}}]{Daniel Dennert}
received his B.Sc. in Computer Science from TU Darmstadt, Germany, where he completed his bachelor’s thesis at the Institute for Intelligent Autonomous Systems~(IAS) TU Darmstadt under the supervision of M.Sc. Daniel Palenicek and Prof. Jan Peters.
He has since started his M.Sc. studies in  Autonomous Systems at the Technical University of Darmstadt.
\end{IEEEbiography}
\begin{IEEEbiography}[{\includegraphics[width=1in,height=1.25in,clip,keepaspectratio]{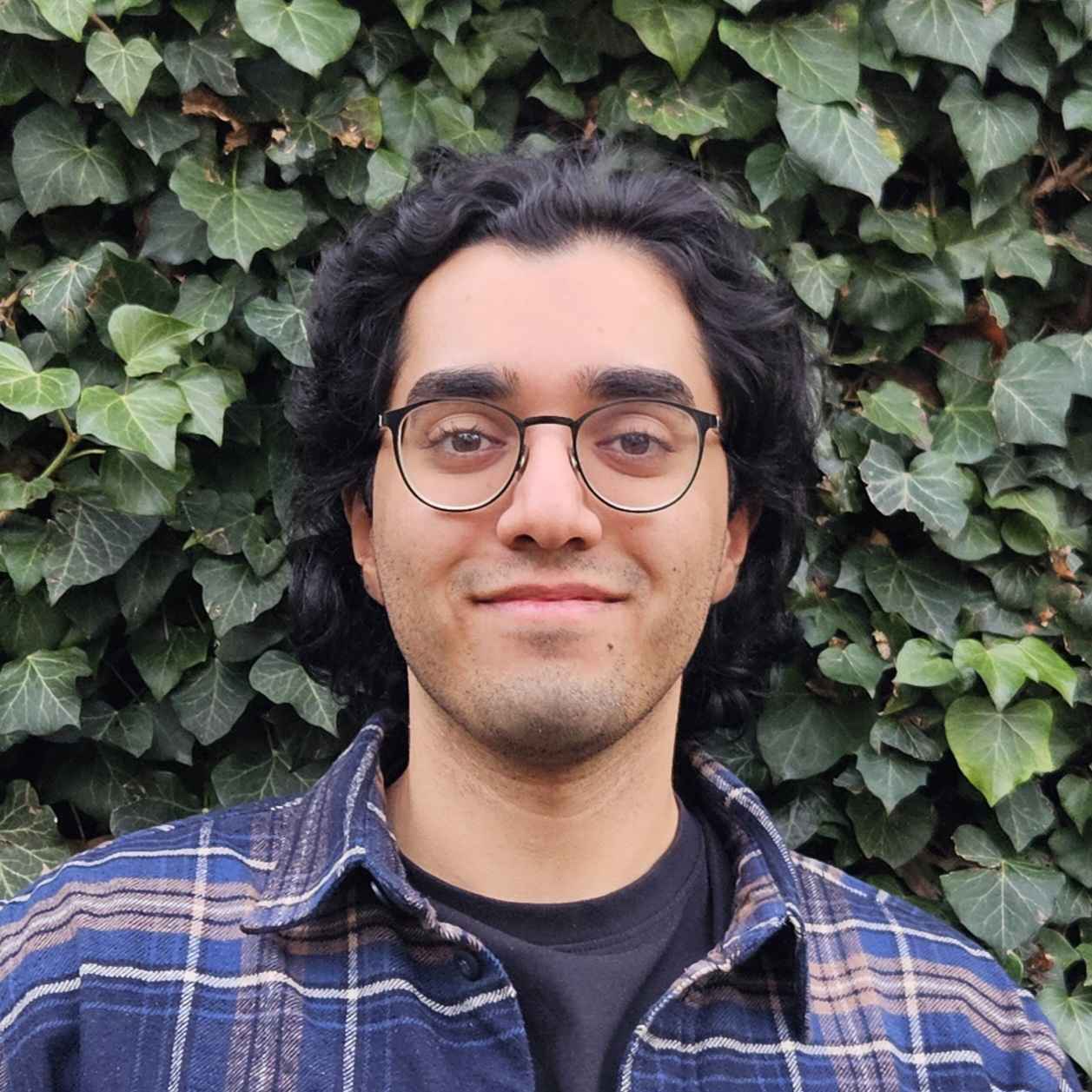}}]{Faran Ahmad}
received his B.Sc. in Computer Science from TU Darmstadt, Germany, where he completed his bachelor’s thesis at the Institute for Intelligent Autonomous Systems~(IAS) TU Darmstadt under the supervision of M.Sc. Daniel Palenicek and Prof. Jan Peters.
He has since started his M.Sc. studies in Computer Science at the Technical University of Darmstadt.
\end{IEEEbiography}
\begin{IEEEbiography}[{\includegraphics[width=1in,height=1.25in,clip,keepaspectratio]{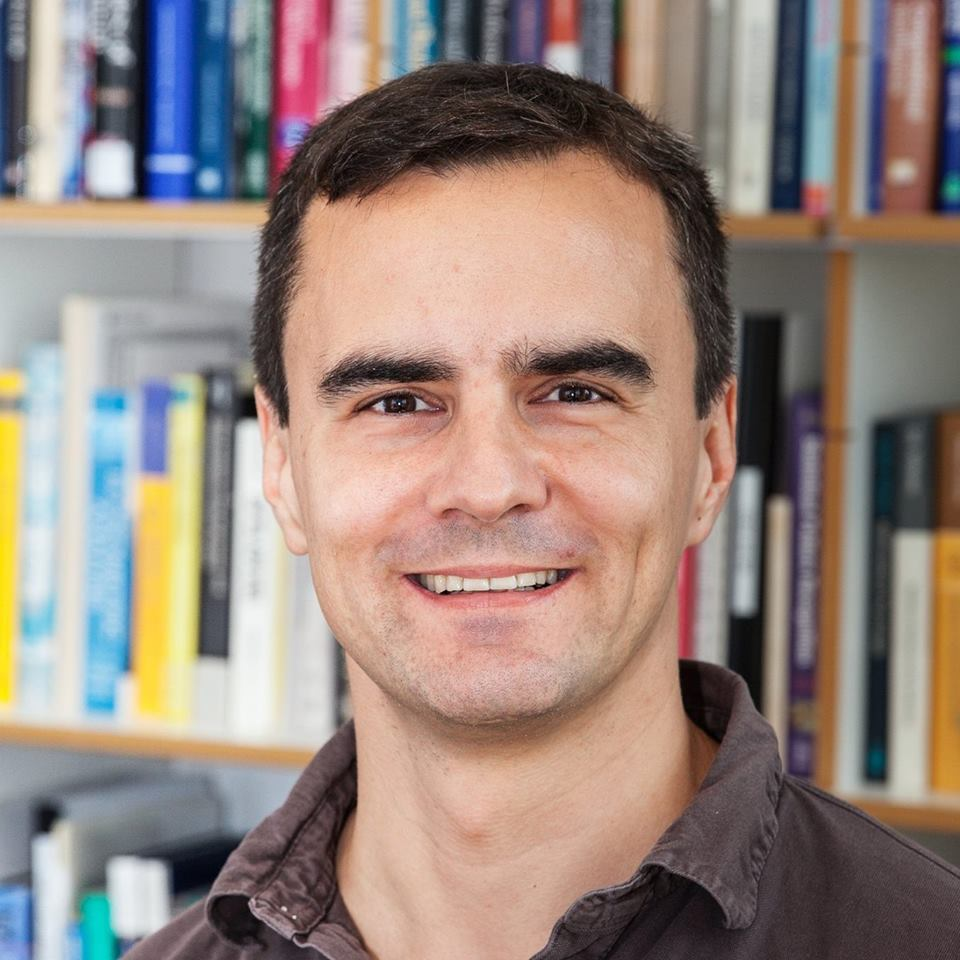}}]{Jan Peters}
(Fellow, IEEE) is a full professor (W3) for Intelligent Autonomous Systems with the Computer Science Department of the Technical University of Darmstadt since 2011 and, he is also the dept head of the research department on Systems AI for Robot Learning (SAIROL) with the German Research Center for Artificial Intelligence (Deutsches Forschungszentrum für Künstliche Intel-
ligenz, DFKI) since 2022.
He is also a founding research faculty member of The Hessian Center for Artificial Intelligence (hessian.AI).
He has received the Dick Volz Best 2007 US PhD Thesis Runner-Up Award, Robotics: Science and Systems - Early Career Spotlight, INNS Young Investigator Award, and IEEE Robotics \& Automation Society’s Early Career Award, as well as numerous best paper awards.
He received an ERC Starting Grant and was appointed an AIAA fellow and ELLIS fellow.
\end{IEEEbiography}
\vfill

{\appendices

\section{From Retrace to On-Policy H-step Value Expansion}
\label{appendix:retrace_to_maxent}
\noindent In this section, we show how \RETRACE{} reduces to the $H$-step value expansion target~(\Eqref{eq:n_step_q}) in the on-policy case, with $\lambda=1$.
First, we perform a simple rewrite from the original \RETRACE{} formulation in~\citet{munos2016retrace} such that it is more in line with the notation in this paper.
Note that in \RETRACE{}, the rollouts are generated with a policy $\mu$. While the value function $V^\pi$ corresponds to the current policy $\pi$.
To make the notation more compact, we will use the following shorthand notations here $\mathcal{R}_t\coloneq\mathcal{R}(\vs_t,\va_t)$, $Q_t\coloneq Q(\vs_t,\va_t)$, $V^\pi_t\coloneq V^\pi(\vs_t)$ and $\mathcal{H}(\vs,\va) \coloneq \alpha \log \pi(\va|\vs)$.
\begin{align*}
    &Q^H_{\text{\textsc{rt}}}(s_0,a_0) = Q_0 + \mathbb{E}_{\va_t\sim\mu} \left[ \sum_{t=0}^H \gamma^t c_t (\mathcal{R}_t + \gamma V^\pi_{t+1} - Q_t) \bigg| \:s_0,\:a_0 \right] \\
      &= Q_0 + \mathbb{E}_{\va_t\sim\mu} \left[ \left[ \sum_{t=0}^H \gamma^t c_t (\mathcal{R}_t - Q_t) \right] + \left[\sum_{t=0}^{H} \gamma^{t+1} c_t V^\pi_{t+1} \right] \bigg| \:s_0,\:a_0  \right] \\
      &= Q_0 + \mathbb{E}_{\va_t\sim\mu} \left[ \left[ \sum_{t=0}^H \gamma^t c_t (\mathcal{R}_t - Q_t) \right] + \left[\sum_{t=1}^{H+1} \gamma^t c_{t-1} V^\pi_t \right] \bigg| \:s_0,\:a_0 \right] \\
      \\
      &= Q_0 + \mathbb{E}_{\va_t\sim\mu} \:\Bigg[ \gamma^0 c_0 (\mathcal{R}_0 - Q_0) \Bigg] \\
      &\qquad\hspace{0.35em}+ \mathbb{E}_{\va_t\sim\mu} \left[ \left[ \sum_{t=1}^H \gamma^t c_t (\mathcal{R}_t - Q_t) \right] + \left[\sum_{t=1}^{H} \gamma^{t} c_{t-1} V^\pi_t \right] \right] \\
      &\qquad\hspace{0.35em}+ \mathbb{E}_{\va_t\sim\mu} \:\Bigg[ \gamma^{H+1} c_{H} V^\pi_{H+1} \Bigg] \\
      \\
      &= Q_0 + \mathcal{R}_0 - Q_0 \\
      &\qquad\hspace{0.35em}+ \mathbb{E}_{\va_t\sim\mu} \left[ \sum_{t=1}^H \gamma^t c_t (\mathcal{R}_t - Q_t) + \gamma^t c_{t-1} V^\pi_t + \gamma^{H+1} c_{H} V^\pi_{H+1} \right] \\
      \\
      &= \mathcal{R}_0 + \mathbb{E}_{\va_t\sim\mu} \left[ \sum_{t=1}^H \gamma^t ( c_t \mathcal{R}_t + c_{t-1} V^\pi_t - c_t Q_t)  + \gamma^{H+1} c_{H} V^\pi_{H+1} \right]
\end{align*}

Now, when $\lambda=1$ the importance sampling weights all become $c_t=1$ and we get
\begin{align*}
    &= \mathcal{R}_0 + \mathbb{E}_{\va_t\sim\mu} \left[ \sum_{t=1}^H \gamma^t (\mathcal{R}_t + V^\pi_t - Q_t) + \gamma^{H+1} V^\pi_{H+1} \right] \\
    \\
    &= \mathcal{R}_0 + \mathbb{E}_{\va_t\sim\mu} \left[ \sum_{t=1}^H \gamma^t (\mathcal{R}_t + \mathbb{E}_{\hat{\va}_t\sim\pi}[Q(\vs_t,\hat{\va}_t) -\mathcal{H}(\hat{\va}_t,\vs_t)] - Q_t) \right] \\
    &\qquad\hspace{0.35em} + \gamma^{H+1} V^\pi(s_{H+1}) \\
\end{align*}

\begin{align*}
    &\quad= \mathcal{R}_0 + \sum_{t=1}^H \gamma^t \Big(\mathbb{E}_{\va_t\sim\mu}[\mathcal{R}_t] + \mathbb{E}_{\hat{\va}_t\sim\pi}[Q(\vs_t,\hat{\va}_t)] \\
    & \quad\qquad\qquad\qquad\hspace{0.35em} - \mathbb{E}_{\hat{\va}_t\sim\pi}[\mathcal{H}(\hat{\va}_t,\vs_t)] - \mathbb{E}_{\va_t\sim\mu}[Q_t] \Big) \\
    & \quad\qquad\hspace{0.35em} + \gamma^{H+1} V^\pi(s_{H+1})
\end{align*}
Since $\mu=\pi$, the expectations over $Q$ are identical and it further reduces to
\begin{align*}
    &\quad= \mathcal{R}_0 + \sum_{t=1}^H \gamma^t \Big(\mathbb{E}_{\va_t\sim\pi}[\mathcal{R}_t] - \mathbb{E}_{\va_t\sim\pi}[\mathcal{H}(\va_t,\vs_t)]\Big) + \gamma^{H+1} V^\pi_{H+1} \\
    &\quad= \mathcal{R}_0 + \mathbb{E}_{\va_t\sim\pi} \left[ \sum_{t=1}^H \gamma^t (\mathcal{R}_t - \mathcal{H}(\va_t,\vs_t)) + \gamma^{H+1} V_{H+1} \right].
\end{align*}
Finally, we see that in the on-policy case with $\lambda=1$, the \RETRACE{} target reduces to the regular \SAC{} $H$-step value expansion target from~\Eqref{eq:n_step_q}.

\newpage
\section{Experiment Details}
\label{sec:experimental_details}
\begin{table*}[hb!]
    \centering
    \caption{Hyperparameters used for \SAC{} and \DDPG{} training. Note, \DDPG{} does not use the $\alpha$ learning rate due to its deterministic policy.}
    \begin{tabular}{ p{0.15\linewidth}|p{0.15\linewidth}|p{0.15\linewidth}|p{0.15\linewidth}|p{0.15\linewidth}|p{0.15\linewidth}  }
     \toprule
     \multirow{2}{*}{\textbf{Parameter}} & \multicolumn{1}{c|}{\textbf{Inverted}} & \multicolumn{1}{c|}{\multirow{2}{*}{\textbf{Cartpole}}} & \multicolumn{1}{c|}{\multirow{2}{*}{\textbf{Hopper}}} & \multicolumn{1}{c|}{\multirow{2}{*}{\textbf{Walker2d}}} & \multicolumn{1}{c}{\multirow{2}{*}{\textbf{HalfCheetah}}} \\
                                & \multicolumn{1}{c|}{\textbf{Pendulum}} &  &  &  &  \\
    \midrule\midrule
     policy learning rate & \multicolumn{5}{c}{$3e^{-4}$} \\ \midrule
     critic learning rate & \multicolumn{5}{c}{$3e^{-4}$} \\ \midrule
     $\alpha$ learning rate & \multicolumn{5}{c}{$5e^{-5}$} \\ \midrule
     target network $\tau$ & \multicolumn{5}{c}{0.005} \\ \midrule
     \RETRACE{} $\lambda$ & \multicolumn{5}{c}{1} \\ \midrule
     number parallel envs & \multicolumn{2}{c|}{1} & \multicolumn{3}{c}{128} \\ \midrule
     min replay size & \multicolumn{2}{c|}{512} & \multicolumn{3}{c}{$2^{16}$} \\ \midrule
     minibatch size & \multicolumn{4}{c|}{256} & \multicolumn{1}{c}{512} \\ \midrule
     discount $\gamma$ & \multicolumn{1}{c|}{0.95} & \multicolumn{3}{c|}{0.99} & \multicolumn{1}{c}{0.95} \\ \midrule
     action repeat & \multicolumn{1}{c|}{4} & \multicolumn{3}{|c|}{2} & \multicolumn{1}{c}{1} \\
    \bottomrule
    \end{tabular}
    \label{tab:hyper_parameters}
\end{table*}

\begin{table*}[hb!]
    \centering
    \caption{Computation times of the training with oracle dynamics. The computational overhead dominates for short horizons already.}
    \begin{tabular}{ p{0.21\linewidth}|p{0.08\linewidth}|p{0.08\linewidth}|p{0.08\linewidth}|p{0.08\linewidth}|p{0.08\linewidth}|p{0.08\linewidth}|p{0.08\linewidth}  }
     \toprule
     \textbf{H} & \textbf{0} & \textbf{1} & \textbf{3} & \textbf{5} & \textbf{10} & \textbf{20} & \textbf{30} \\
     \midrule\midrule
     \textbf{InvertedPendulum} \\
     \SAC{}-\CExpansion{}  & 6m &  8m & 10m & 12m & 18m & 28m & 42m \\
     \SAC{}-\AExpansion{}  & 6m & 14m & 30m & 42m & 1h  & 2h  & 3h \\
     \RETRACE{} & 6m &  6m &  7m &  8m & 10m & 14m & 18m \\
     \midrule
     \textbf{Hopper \& Walker2d} \\
     \SAC{}-\CExpansion{}  & 20m & 1h20m  & 2h30m  & 5h30m  & 10h30m  & 20h20m  & 30h20m \\
     \SAC{}-\AExpansion{}  & 20m & 9h 40m & 35h    & 55h    & 114h    & 167h    & 278h \\
     \RETRACE{} & 20m & 20m    & 22m    & 23m    & 27m     & 30m     &  40m   \\
     \midrule
     \textbf{HalfCheetah} \\
     \SAC{}-\CExpansion{}  & 40m & 3h15m & 5h 40m & 13h  & 25h  & 49h & 77h \\
     \SAC{}-\AExpansion{}  & 40m & 30h   & 62h    & 100h & 168h &  -  &  -  \\
     \RETRACE{} & 40m & 41m   & 45m    &  50m &  52m &  1h & 1h30m\\
    \bottomrule
    \end{tabular}
    \label{tab:computation_times}
\end{table*}
\noindent This section provides detailed information regarding policy and $Q$-function networks, as well as the learned dynamics model.
Further, we include hyperparameters of the training procedure.

\subsubsection{Policy Representation}
The policy is represented by a neural network.
It consists of two hidden layers with $256$ neurons each and ReLU activations.
The network outputs a mean vector and a per-dimension standard deviation vector which are then used to parameterize a diagonal multivariate Gaussian.

\subsubsection{Q-function Representation}
To represent the $Q$-function we use a double $Q$ network.
Each network has two hidden layers with $256$ neurons each and ReLU activations and outputs a single scalar $Q$ value.
The minimum of the two $Q$ networks is taken in critic and actor targets.

\subsubsection{Learned Dynamics Model}
The learned dynamics model is a reimplementation of the one introduced in~\citet{janner2019mbpo}.
It is an ensemble of $5$ neural networks with $4$ hidden layers with $256$ neurons each and ReLU activations.
To generate a prediction, one of the networks from the ensemble is sampled uniformly at random.
The selected network then predicts mean and log variance over the change in state.
This change is added to the current state to get the prediction for the next state.

\subsubsection{Hyperparameters for Training}
For a fair comparison, hyperparameters are shared across \SAC{} and \DDPG{} for the different expansion methods.
In Table~\ref{tab:hyper_parameters} we report the hyperparameters across all experiments.

\subsubsection{Computation Times}
We report some of the computation times for the reader to get a better grasp of the computational complexity.
Even though our GPU-only implementation using Brax~\cite{brax2021github} is much more efficient than having a CPU-based simulator and transferring data back and forth to the GPU, the computational complexity of rolling out the physics simulator in the inner training loop, is still significant.
This relates to complexity in computation time as well as VRAM requirements.
Table~\ref{tab:computation_times} shows the computation times on an NVIDIA A100 GPU.
\newpage

\begin{figure*}[ht!]
    \centering
    \includegraphics[width=\linewidth]{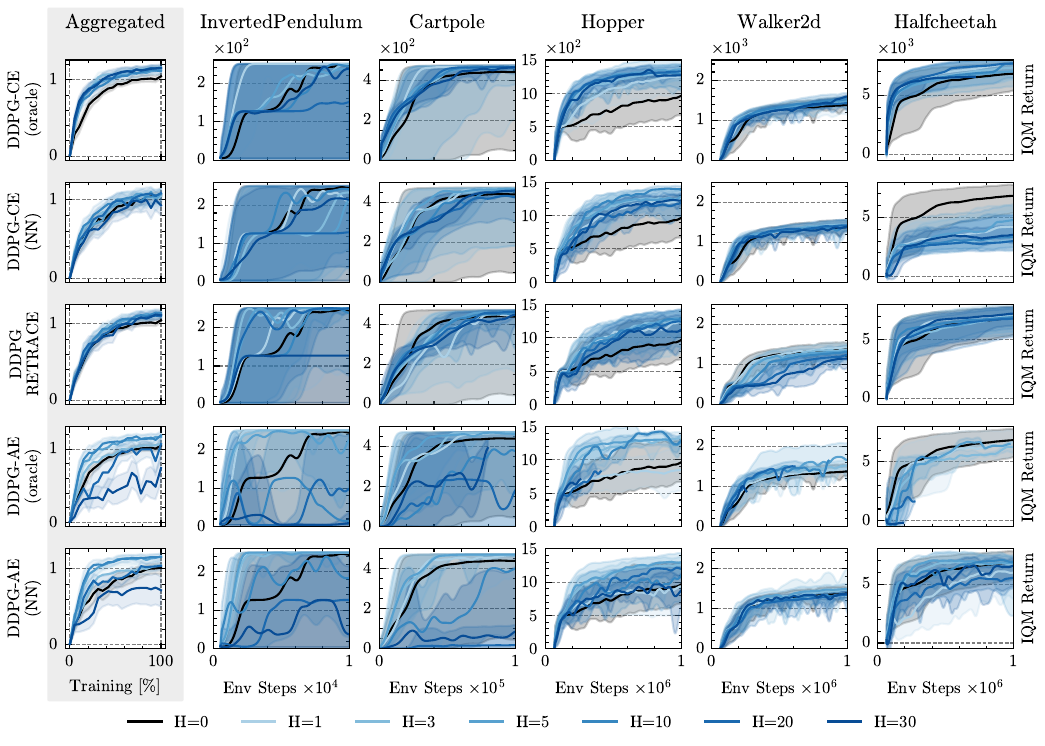}
    \caption{
    \textbf{DDPG Experiments on continuous control tasks.}
    Shows the diminishing return of \CExpansion{} and \AExpansion{} methods using multiple rollout horizons $H$ for a \DDPG{} agent.
    \CExpansion{} does not benefit noticeably from horizons larger than 5 in most cases.
    On average, \CExpansion{} can slightly benefit from oracle dynamics over our learned model, while for \AExpansion{}, this is not the case.
    Looking at individual runs, often oracle and learned dynamics models perform very similarly.
    Model-free \RETRACE{} performs comparably to the learned model and even more stably with regard to longer rollout horizons.
    For \AExpansion{} larger horizons can be detrimental, even with an oracle model.
    We plot the \IQM{} episode undiscounted return (solid line) and $90\%$ \IPR{} (shaded area) against the number of real environment interaction steps for $9$ random seeds.
    Some \AExpansion{} runs for larger $H$ terminate early due to exploding gradients (See~\Secref{sec:gradient_analysis}).
    }
    \label{fig:ddpg_all}
\end{figure*}
\section{Individual DDPG Experiments}
\noindent \Figref{fig:ddpg_all} shows all \DDPG{} runs for all expansion methods individually as well as aggregated for the continuous control enviornments.
It is the equivalent of \Figref{fig:sac_mve_ivg_retrace} from the main paper.
Qualitatively, the results and findings are the same for \DDPG{} as they were for \SAC{}.

\newpage
\section{Gradient Experiments}
\noindent Figures~\ref{fig:critic_grads_all}, \ref{fig:actor_grads_all}, \ref{fig:critic_grads_all_ddpg} and~\ref{fig:actor_grads_all_ddpg} show critic and actor gradients' mean and standard deviation for the different expansion methods of \SAC{} and \DDPG{}, respectively.

\begin{figure*}[ht!]
    \centering
    \includegraphics[width=\linewidth]{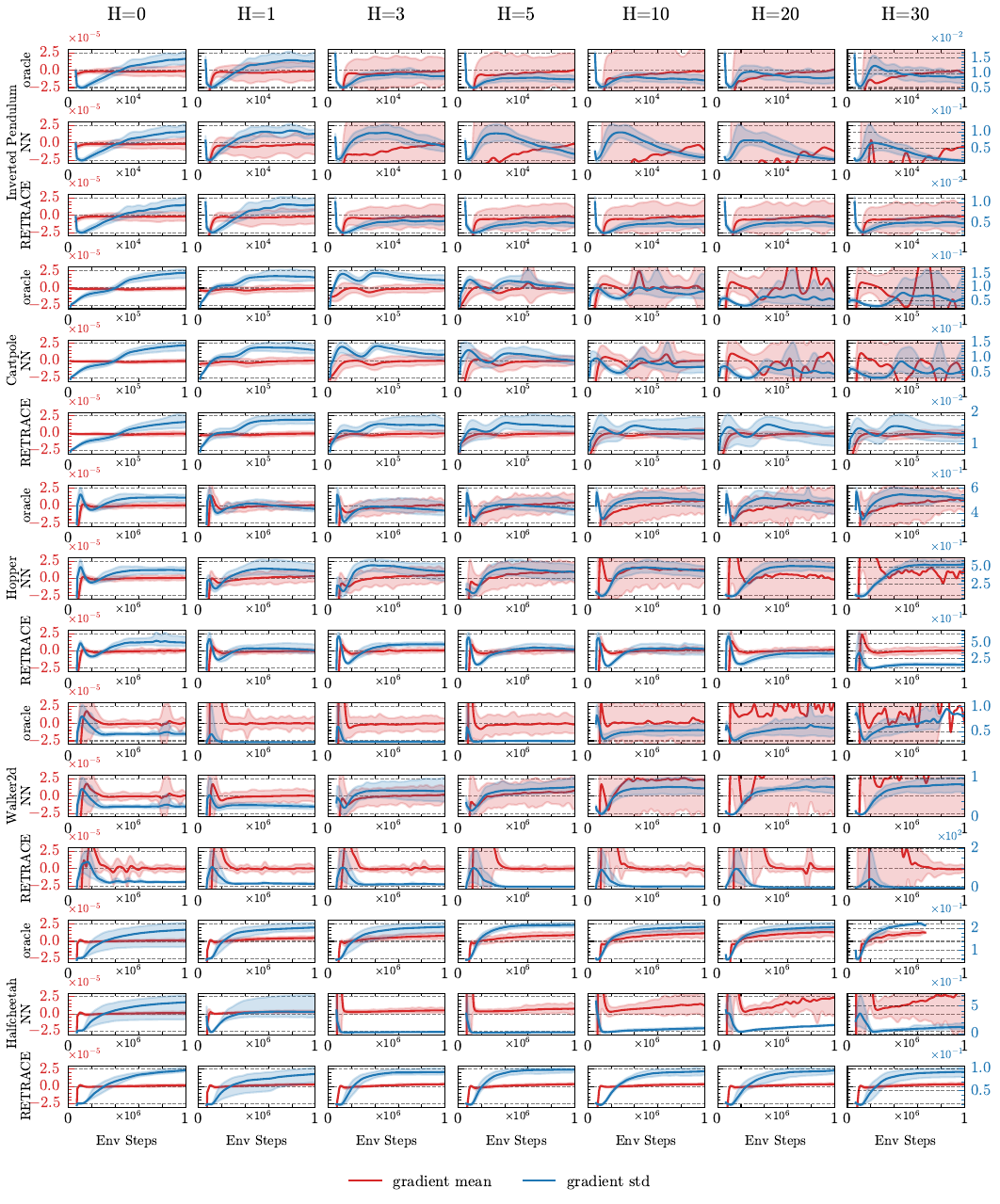}
    \caption{
    \textbf{SAC Critic Expansion Gradients.}
    These plots show the critic's gradients mean and standard deviation of \SAC{}-\CExpansion{} for all environments and all rollout horizons.
    The values are in reasonable orders of magnitude and, therefore, the diminishing returns in \SAC{}-\CExpansion{} cannot be explained via the critic gradients.
    }
    \label{fig:critic_grads_all}
\end{figure*}

\begin{figure*}[ht!]
    \centering
    \includegraphics[width=\linewidth]{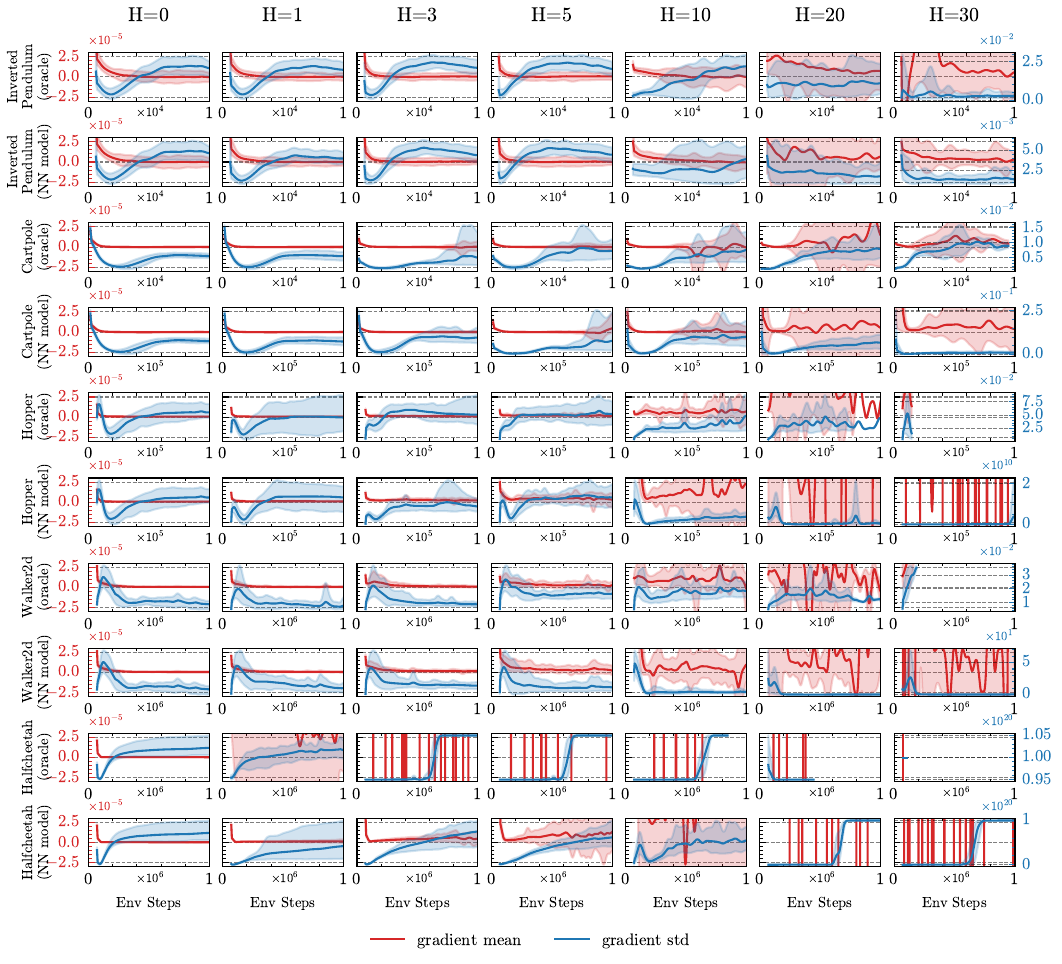}
    \caption{
    \textbf{SAC Actor Expansion Gradients.}
    These plots show the actor's gradients mean and standard deviation of \SAC{}-\AExpansion{} for all environments and all rollout horizons.
    Even though the gradients' mean have reasonable values (close to zero), except for the Halfcheetah environment, the standard deviation shows spikes and abnormal values for increasing rollout horizons, e.g., note the spikes in the standard deviation in the Hopper (NN model) experiments for horizons $20$ and $30$ (row 6, last two columns), which go up to the clipped value of $10^2$.
    These increase in variance can be correlated with the lower return in Figure~\ref{fig:sac_mve_ivg_retrace} (row 5, col 3).
    }
    \label{fig:actor_grads_all}
\end{figure*}

\begin{figure*}[ht!]
    \centering
    \includegraphics[width=\linewidth]{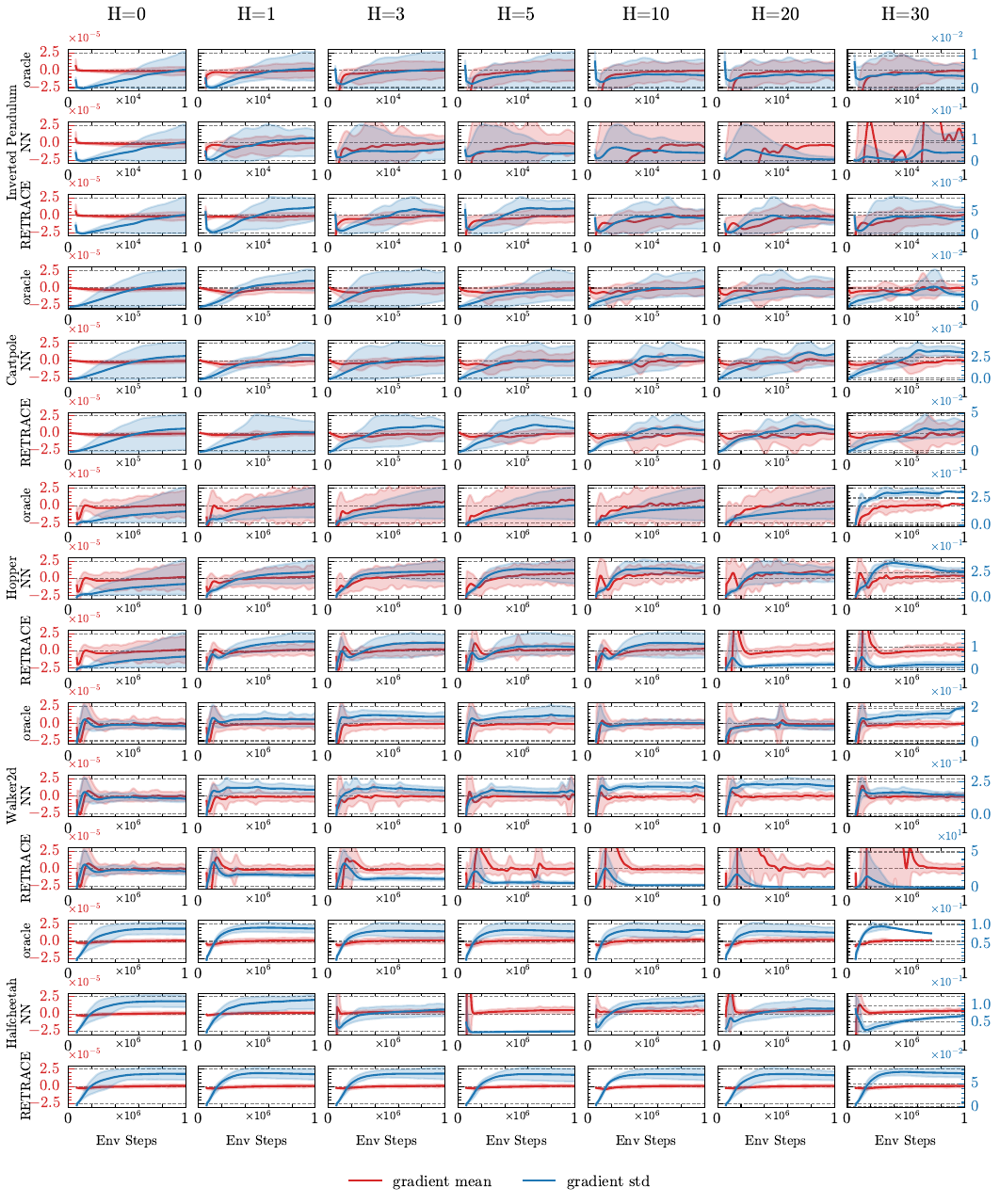}
    \caption{
    \textbf{DDPG Critic Expansion Gradients.}
    These plots show the critic's gradients mean and standard deviation of \DDPG{}-\CExpansion{} for all environments and all rollout horizons.
    The values are in reasonable orders of magnitude and, therefore, the diminishing returns in \DDPG{}-\CExpansion{} cannot be explained via the critic gradients.
    }
    \label{fig:critic_grads_all_ddpg}
\end{figure*}

\begin{figure*}[ht!]
    \centering
    \includegraphics[width=\linewidth]{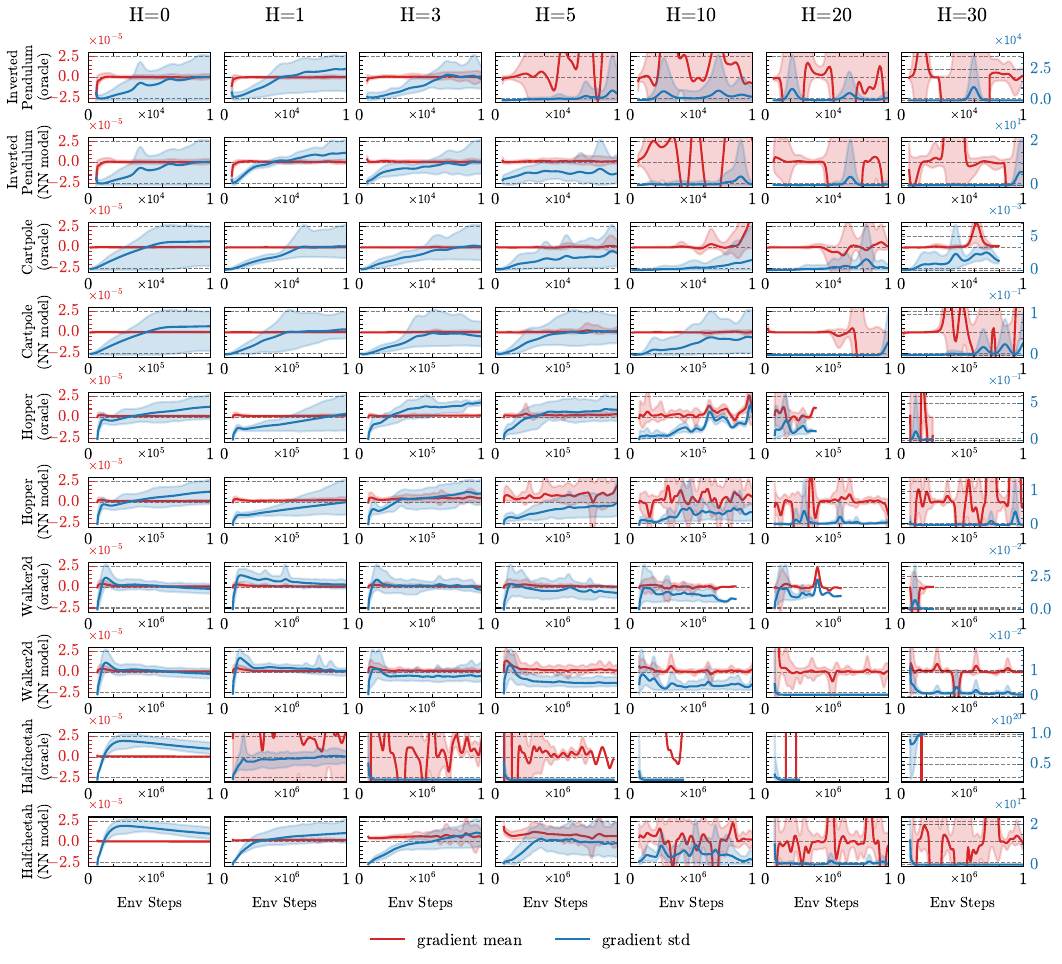}
    \caption{
    \textbf{DDPG Actor Expansion Gradients.}
    These plots show the actor's gradients mean and standard deviation of \DDPG{}-\AExpansion{} for all environments and all rollout horizons.
    }
    \label{fig:actor_grads_all_ddpg}
\end{figure*}

}

\end{document}